%% file: egpaper_for_review_camera_ready.tex
\documentclass[10pt,twocolumn,letterpaper]{article}

\usepackage{iccv}
\usepackage{times}
\usepackage{epsfig}
\usepackage{graphicx}
\usepackage{amsmath}
\usepackage{amssymb}

\usepackage{adjustbox, color, paralist, enumerate, booktabs, multirow, array, tabu, float, bm}

\usepackage{comment}

\usepackage{xcolor}
\usepackage{enumitem}
\usepackage{colortbl}
\definecolor{sh_gray}{rgb}{0.84,0.84,0.84}
\definecolor{sh_gray2}{rgb}{1,0.89,0.75}
\definecolor{color3}{rgb}{0.95,0.95,0.95}
\definecolor{color4}{rgb}{0.96,0.96,0.86}
\definecolor{color5}{rgb}{0.90,0.90,0.90}
\usepackage{tabularx}

\usepackage[switch]{lineno}  %

\newcommand{\widthscalefive}{0.08}
\newcommand{\widthscalesix}{0.12}

\newcommand{\widthscalefirst}{0.17}
\newcommand{\widthscaleblur}{0.11}

\usepackage[pagebackref=true,breaklinks=true,letterpaper=true,colorlinks,bookmarks=false]{hyperref}

\def\ourmethod{SPAIR}

\iccvfinalcopy 

\def\iccvPaperID{9231} 

\ificcvfinal\fi

\begin{document}

\title{Spatially-Adaptive Image Restoration using Distortion-Guided Networks}

\author{Kuldeep Purohit$^{1}$ \qquad Maitreya Suin$^{2}$ \qquad A. N. Rajagopalan$^{2}$ \qquad Vishnu Naresh Boddeti$^{1}$\\
$^1$ Michigan State University \hspace{1cm}
$^2$ Indian Institute of Technology Madras\\
} 

\maketitle
\ificcvfinal\thispagestyle{empty}\fi

\input{paper_parts/abstract_cameraready}
\section{Introduction}
\input{paper_parts/introduction2}

\vspace{-1mm}
\section{Related Works}
\input{paper_parts/related_works2_cameraready}

\vspace{-1mm}

\section{Proposed Network Architecture}
\label{method}
\input{paper_parts/method2_cameraready}

\section{Datasets and Implementation Details}
\label{Implementation}

\noindent \textbf{Rain-Streaks:} 
Using the same experimental setups of the recent approaches on image deraining~\cite{mspfn2020}, we train our model on $13$,$712$ clean-rain image pairs gathered from multiple datasets~\cite{fu2017removing,li2016rain,yang2017deep,zhang2018density,zhang2019image}. 
With this single trained model, we perform evaluation on different test sets, including Rain100H~\cite{yang2017deep}, Rain100L~\cite{yang2017deep}, Test100~\cite{zhang2019image}, Test2800~\cite{fu2017removing}, and Test1200~\cite{zhang2018density}. We also report the error reduction error for each method relative to the best method by translating PSNR to RMSE ($\textrm{RMSE} \propto \sqrt{10^{-\textrm{PSNR}/10}}$) and SSIM to DSSIM ($\textrm{DSSIM} = (1 - \textrm{SSIM})/2$). We also evaluate \ourmethod{} on SPANet Dataset \cite{wang2019spatial} (real-world rain) containing $2\times10^5$ training and $1000$ testing images.

\noindent \textbf{Raindrop:} We use the AGAN dataset \cite{qian2018attentive} with 861 training and $58$ test samples. Images were generated by placing a raindrop covered glass between the camera and scene. 

\noindent \textbf{Shadow:}  We evaluate our model using a challenging benchmark ISTD~\cite{wang2018stacked} containing $1300$ (train) and $540$ (test) images (with real shadows and diverse textured scenes).

\noindent \textbf{Motion Blur:} 
We follow the configuration of \cite{Maitreya2020, dmphn2019, deblurganv2, tao2018scale} and use the GoPro \cite{gopro2017} dataset containing $2$,$103$ image pairs for training and $1$,$111$ pairs for evaluation. Furthermore, to demonstrate generalizability, we directly evaluate our GoPro trained model on the test set of  HIDE~\cite{shen2019human} and RealBlur~\cite{rim_2020_realblur} datasets. The HIDE dataset is specifically collected for human-aware motion deblurring, containing $2$,$025$ test images. While the GoPro and HIDE datasets are generated by averaging real videos, the blurred images in RealBlur-J dataset are captured in real-world conditions.

\noindent\textbf{Implementation Details:} The $Net_R$ for each degradation is trained to minimize $l_1$ reconstruction loss between the  output and the GT clean image. $Net_L$ is trained using binary cross entropy loss with respect to the GT binary mask. Each training batch contains randomly cropped RGB patches of size $256\times256$ from degraded images that are randomly flipped horizontally or vertically. The batch-size was $8$ for rain-streak, raindrop, and shadow removal and $16$ for deblurring. Both networks use Adam optimizer with initial leaning rate $2 \times 10^{-4}$, halved after every $50$ epochs. We use PyTorch library and RTX 2080Ti GPU.

\section{Experimental Evaluation}
\label{Experimental_comparisons}
\input{paper_parts/rainstreak}

\input{paper_parts/raindrop}

\input{paper_parts/shadow}

\input{paper_parts/blur_cameraready}

\section{Network Analysis}
\label{ablation2}
\input{paper_parts/ablation2}

\section{Conclusions}
\input{paper_parts/conclusion}

{\small
\bibliographystyle{ieee_fullname}
\bibliography{egbib}
}

\end{document}

%% file: paper_parts/abstract_cameraready.tex
\begin{abstract}

We present a general learning-based solution for restoring  images  suffering  from  spatially-varying  degradations. Prior approaches are typically degradation-specific and employ the same processing across different images and different pixels  within.   However,  we  hypothesize  that  such  spatially rigid processing is suboptimal for simultaneously restoring the degraded pixels as well as reconstructing the clean regions of the image. To overcome this limitation, we propose \ourmethod{}, a network design that harnesses distortion-localization information and dynamically adjusts computation to difficult regions in the image. \ourmethod{} comprises of two components, (1) a localization network that identifies degraded pixels,  and (2) a restoration network that exploits knowledge from the localization network in filter and feature domain to selectively and adaptively restore degraded pixels. Our key idea is to exploit the non-uniformity of heavy degradations in spatial-domain and suitably embed this knowledge within distortion-guided modules performing  sparse normalization, feature extraction and attention. Our architecture is agnostic to physical formation model and generalizes across several types of spatially-varying degradations. We demonstrate the efficacy of \ourmethod{} individually on four restoration tasks- removal of rain-streaks, raindrops, shadows and motion blur. Extensive qualitative and quantitative comparisons with prior art on $11$ benchmark datasets demonstrate that our degradation-agnostic network design offers significant performance gains over state-of-the-art degradation-specific architectures. Code available at https://github.com/human-analysis/spatially-adaptive-image-restoration.
\end{abstract}

%% file: paper_parts/introduction2.tex
\begin{figure}[t]
  \centering 
  \setlength{\tabcolsep}{2pt}
  \begin{tabular}{cccc}
  \includegraphics[width=.218\linewidth,height = .14\linewidth]{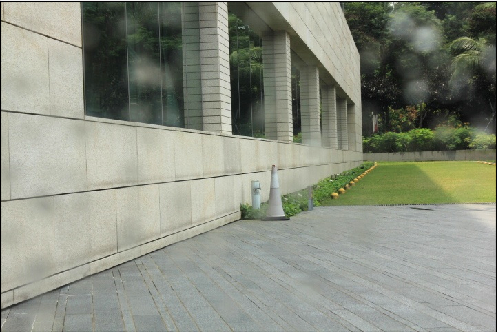}&
  \includegraphics[width=.218\linewidth,height = .14\linewidth]{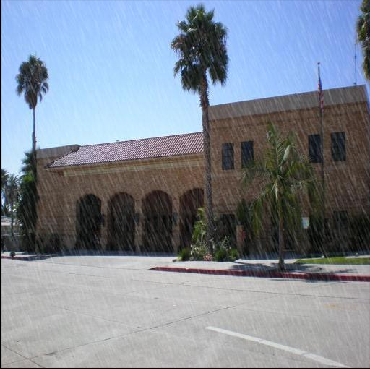}&
  \includegraphics[width=.218\linewidth,height = .14\linewidth]{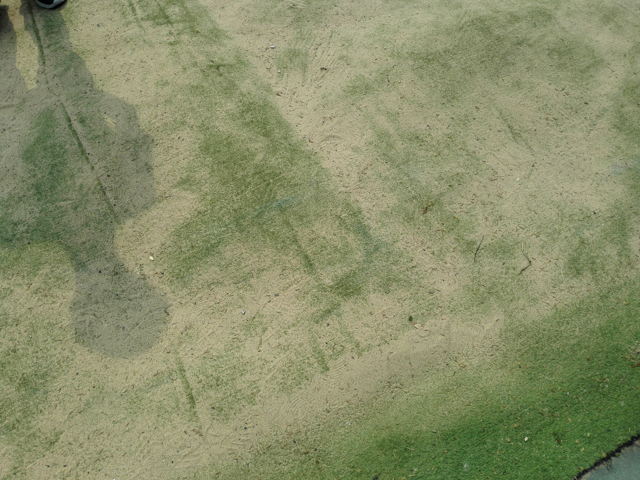}&
  \includegraphics[width=.218\linewidth,height = .14\linewidth]{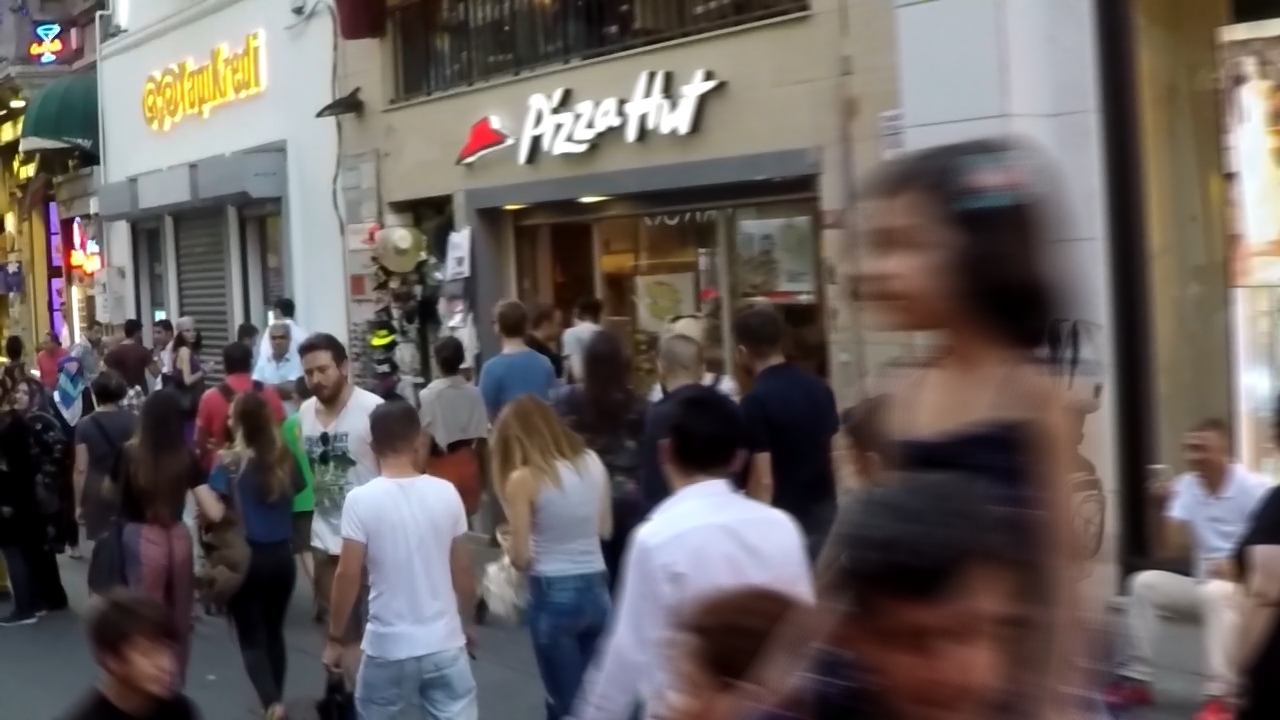}
  \\
  \includegraphics[width=.218\linewidth,height = .14\linewidth]{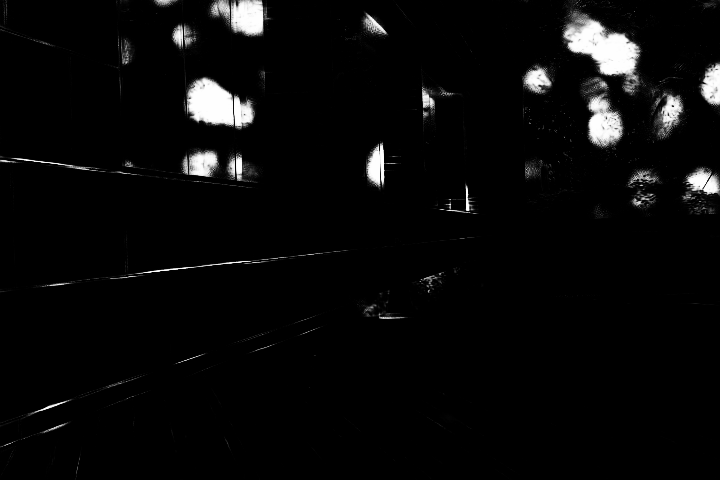}&
  \includegraphics[width=.218\linewidth,height = .14\linewidth]{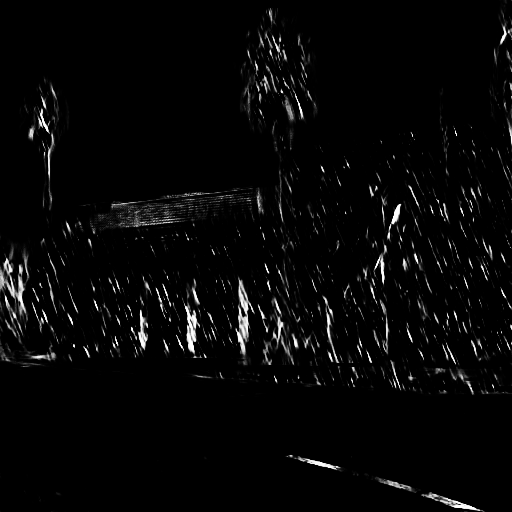}&
  \includegraphics[width=.218\linewidth,height = .14\linewidth]{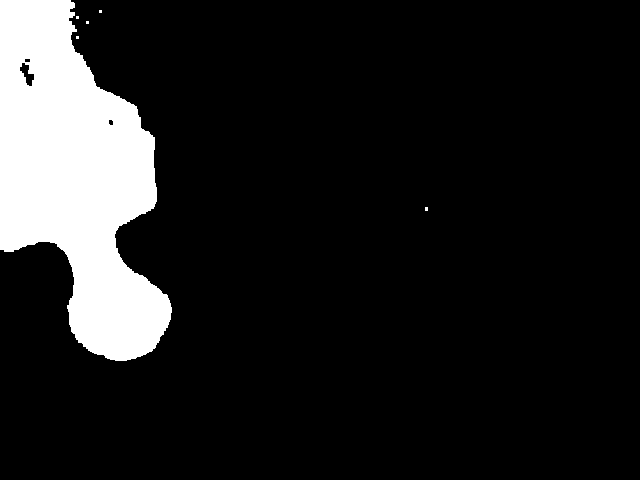}&
  \includegraphics[width=.218\linewidth,height = .14\linewidth]{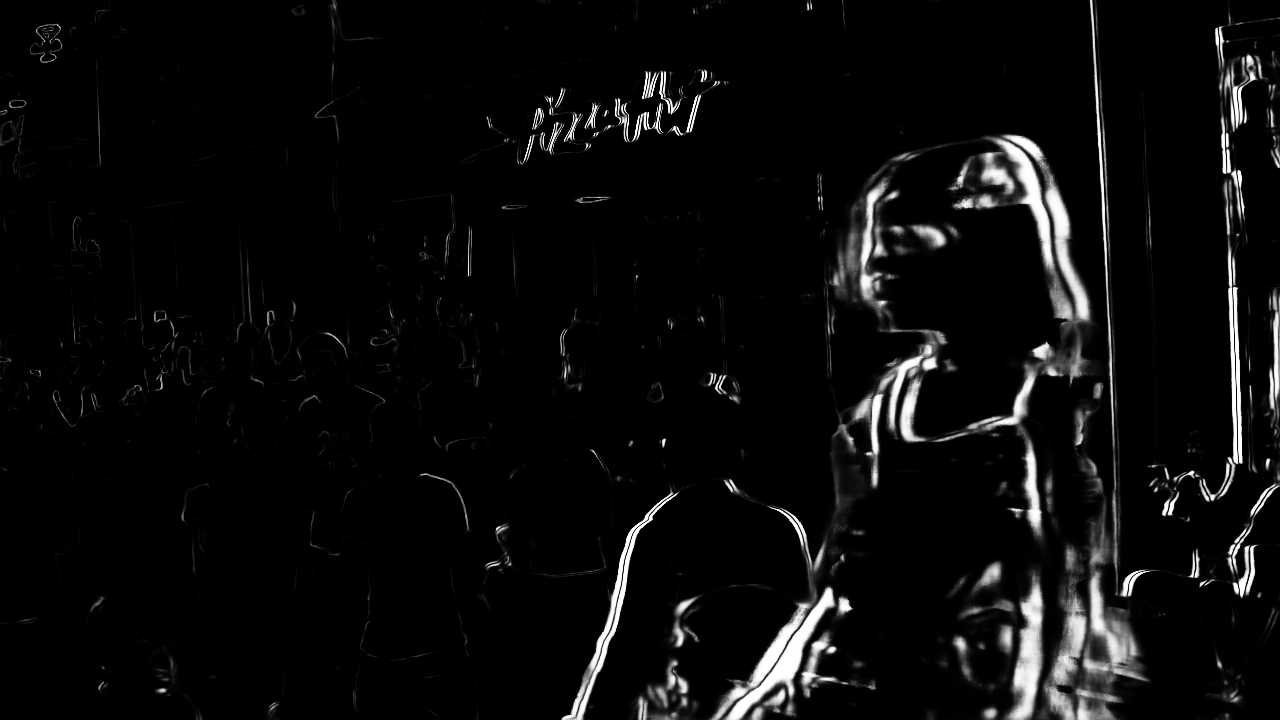}
  \\
  \end{tabular}
\caption{Visualization of degradation masks. The two rows show degraded input images and corresponding predicted masks.\label{fig:representative}}
\vspace{-0.5cm}
\end{figure}

Images are often degraded during the data acquisition process, especially under non-ideal imaging conditions. Such degradations can be attributed to the medium and dynamics between the camera, scene elements and the illumination. For instance, as shown in Fig. \ref{fig:representative} , (1) precipitation leads to snow/rain streaks occupying the volume between the scene and the camera, (2) presence of rain-drops on the camera lens causes significant degradation in scene visibility, (3) relative motion between the camera or scene elements results in motion blur, and (4) harsh illumination conditions can induce harsh shadows. Despite the disparate source of degradations, they share the same underlying motif that affects the image quality, namely, degradation that is spatially-varying in nature. For example, raindrops and shadows degrade monolithic parts of the image depending on their size and location, motion blur varies with scene depth and degree of motion, and rain streaks effect only sparse regions whose orientation depends on the relative rain direction. Fig. \ref{fig:representative} shows representative examples of degraded images and respective distortion-maps. It can be seen that a large number of pixels undergo little or no distortion. Another observation is that the amount of distortion and its spatial distribution is different in every image.

Restoring such images is vital to improve their aesthetic quality as well as the performance of downstream tasks, viz, detection, segmentation, classification, and tracking. Convolutional neural networks (CNNs) currently typify the state-of-the-art for various image restoration tasks. Despite recent progress, existing approaches share several key limitations. Firstly, all layers in their networks are generic CNN layers, which apply the same set of spatially-invariant filters to every degraded image. Such layers are limited in their ability to invert degradations that are highly image-dependent and spatially-varying. Secondly, most network architectures are specifically tailored for individual degradation types as they are based on image formation models. Thirdly, the distortion-localization information embedded in the labeled datasets remains unused or sub-optimally used in all existing solutions.

Static CNN based models trained to directly regress clean intensities from degraded ones, perform poorly when input contains unaffected regions as well as severe intensity distortions in different spatial regions. Conceptually, a stack of fixed learned filters, excelling at restoring pixels degraded with large distortions might not be suitable for reconstructing the texture from unaffected regions. Practically, we observe that such designs often yield poor reconstruction performance (introduce unwanted changes or artifacts on pixels that are not degraded in the input to begin with). The image-dependent nature of spatial distribution and magnitude of distortions only exacerbates the problem faced by static CNNs.

Motivated from the understanding that a restoration network can benefit from adapting to the degradation present in each test image, we propose a distortion-aware model to simultaneously realize the twin goals of restoration and reconstruction. Our spatially-adaptive image restoration architecture (referred to as \ourmethod{}) is suited for any type of degradation which selectively affects parts of the image. It comprises of two components- a distortion-localization network ($Net_L$) and a spatially-guided restoration network ($Net_R$). $Net_L$ gathers information from the entire image to estimate a binary mask (localizing high intensity distortions) which steers the processing in $Net_R$ to selectively improve only degraded regions. 

The proposed $Net_R$ comprises of 3 distortion-guided blocks- spatial feature modulator (SFM), sparse convolution module (SC) and a custom sparse non-local module (SNL). SFM utilizes the output mask and intermediate features from $Net_L$ to modulate the feature statistics of intermediate features in $Net_R$. SC and SNL improve features in the spatially-sparse degraded regions in an image-dependent manner, without affecting the features in clean regions. SNL locally restores features in distorted regions by adaptively gathering global context from all clean regions. Our key contributions are:

\noindent \textbf{--} A two-stage framework to systematically exploit distortion-localization knowledge for directly addressing the challenges associated with diverse spatially-varying degradations in an interpretable manner. It achieves the twin goals of restoration and reconstruction and works across diverse degradation-types. 

\noindent \textbf{--} Distortion-guided spatially-varying modulation of features statistics in $Net_R$ with the help of distortion-mask and features from a pretrained $Net_L$.

\noindent \textbf{--} Distortion-guided feature extraction with the help of SC (for local context) and a novel SNL (for global context) modules. These components facilitate spatially-varying restoration while controlling receptive field in an image and location-adaptive manner.  

\noindent \textbf{--} We demonstrate the versatility of \ourmethod{} by setting new state-of-the-art on \textbf{11} synthetic and real-world datasets for various spatially-varying restoration tasks (removing rain-streaks, rain-drops, shadows, and motion blur), outperforming existing approaches designed with task-specific network-engineering. Further, we provide detailed analysis, qualitative results, and generalization tests.

%% file: paper_parts/related_works2_cameraready.tex
\noindent \textbf{Adaptive Inference:} Adaptive inference techniques \cite{wang2018skipnet,teja2018hydranets,graham20183d,li2019improved} have attracted increasing interest since they enable input-dependent alteration of CNN structure. One class of methods dynamically skip subsets of layers in cascaded CNNs during inference \cite{wu2018blockdrop,teja2018hydranets,figurnov2017spatially}. \cite{figurnov2016perforatedcnns} passes sampled pixels (using a random pattern which is fixed during inference) to CNN layers and fills the remaining locations using simple interpolation.
Few approaches \cite{graham2017submanifold,graham20183d} exploit sparsity in the input image itself using sub-manifold sparse convolutions, but are unsuitable for non-sparse input data. 

However, none of these approaches afford the fine-grained spatial-domain control necessary for spatially-varying image restoration at multiple intermediate layers. For instance, the approaches that skip processing of some layers or prune the network still filter the degraded and other image regions with the same parameters. Methods such as \cite{figurnov2017spatially} are only applicable to cascade of consecutive residual layers, and do not generalize to encoder-decoder designs (typically used for image restoration) where conditionally altering network depth or channel width is non-trivial. The arbitrary rejection of spatial-domain information proposed in \cite{figurnov2016perforatedcnns} is ill-fitted for general restorations tasks.

\noindent \textbf{Raindrop Removal:} Solutions for raindrop removal include both classical as well as CNN based approaches.
\cite{kanthan2015rain} proposed a clustering and median filtering based restoration, while CNN based approaches include, shallow CNNs \cite{eigen2013restoring} but with limited performance, a convolutional-LSTM based model for ``joint'' learning of rain-map and rain-free image \cite{qian2018attentive}, and a deeper CNN \cite{DuRN}. \cite{quan2019deep} instead leveraged physical models of raindrop properties (including closedness and roundness) to estimate drop-probability. In contrast to these methods, \ourmethod{} advocates for a pixel selective and adaptive processing to remove raindrops.

\noindent \textbf{Rain-streak Removal:} Conventional deraining methods~\cite{ding2016single,zhu2017joint,li2016rain,luo2015removing} adopt a model-driven methodology utilizing physical properties of rain and prior knowledge of background scenes into an optimization problem. CNN-based approaches include end-to-end negative residual mapping \cite{fu2017removing}, deeper CNN \cite{yang2017deep}, multi-stage CNNs with recurrent connections \cite{li2018recurrent}, CNN for predicting density (heavy, medium, light) during deraining \cite{zhang2018density}, concatenating rain-map for deraining \cite{yang2019joint}. However, layers in these approaches process all image regions with the same filters (without pixel adaptation). \cite{wang2020model} presents  model-driven CNN with convolutional dictionary learning. Wang et.al. \cite{wang2019spatial} predicts a rain-map and multiplies it element-wise with feature-maps to enhance them. While \ourmethod{} also utilizes a mask, there are fundamental differences. We estimate a binary mask and utilize it more comprehensively, including for sparse filtering, attention weight calculation and guiding it to non-degraded image regions. 
\ourmethod{} significantly differs from rain-guided models of \cite{yang2019joint,qian2018attentive} in three aspects. (1) They only concatenate the rain-mask at the input. In contrast, we exploit distortion-mask to only perform convolutions and non-local operations on degradation regions. We also transfer feature statistics from clean to degraded regions at multiple intermediate layers using SFM. (2) They lack global context. \ourmethod{} contains SNL module that adaptively gathers all features values within the clean regions of the image. (3) All pixels are passed through same network with spatially rigid processing, which directly contrasts with our work. Ours is the first approach to exploit explicit degradation-guidance to selectively processes degraded pixels and reduce the effect on unaffected regions, for a variety of spatially-sparse degradations.

\noindent \textbf{Shadow Removal:} Early works often erased shadows via user interaction or by transferring illumination from non-shadow regions to shadow regions~\cite{guo2012paired,khan2015automatic}. More robust results have been achieved using CNN based approaches which include using multiple networks \cite{hu2018direction}, DeshadowNet for illumination estimation in shadow regions \cite{qu2017deshadownet}, stacked conditional GANs \cite{wang2018stacked}, ARGAN to detect and remove shadow with multiple steps \cite{ding2019argan},  RIS-GAN \cite{zhang2020ris} to estimate negative residual images and inverse illumination maps for restoration, and finally a cascade of dilated convolutions to jointly estimate shadow-mask and shadow-free image \cite{cun2020towards}. In contrast to the aforementioned approaches, we propose a two-stage framework wherein the distortion-mask and intermediate learned features of $Net_L$ are employed in a principled manner for region-aware and selective restoration.

\noindent \textbf{Motion Blur Removal:} 
Traditional approaches \cite{lai2016comparative} designed priors on image and motion (eg. locally linear blur kernels \cite{sun2015learning,gong2017motion}, planar scenes \cite{purohit2019planar}) but with limited success in general 3D and dynamic blurred scenes. Recent CNN-based methods directly estimate the latent sharp image \cite{nah2017deep}, wherein encoder-decoder designs that aggregate features in a coarse-to-fine manner have been proposed \cite{nah2017deep,tao2018scale,gao2019dynamic,purohit2019bringing}. Additionally, \cite{zhang2018dynamic} explored a design composed of multiple CNNs and RNN and \cite{zhang2019deep} proposed a patch-hierarchical network and stacked its copies along depth to achieve state-of-the-art performance. \cite{mtrnn2020} proposed a recurrent design for efficient
deblurring. A limitation shared by all of these methods is
the absence of spatially varying adaptive layers. \cite{purohit2020region,purohit2019efficient} proposed direction-based feature extraction modules suited for efficient motion deblurring. \cite{Maitreya2020} inserts
adaptive convolution and attention within the layers
of \cite{zhang2019deep} to boost its results. Our distortion-guided sparse architecture performs better than such patch hierachical designs, while generalizing beyond motion-blur and offering consistent gains across other degradations. 

\noindent  \textbf{Architectures for general Restoration}
A few solutions have been proposed in literature to address multiple degradation-types. For instance, DuRN~\cite{DuRN_cvpr19} make task dependent alterations in their network structure. Similarly, OWAN~\cite{suganuma2019attention} was proposed to handle multiple degradations present within the same image. However, \cite{suganuma2019attention} only addresses simple synthetic degradations that are similar in nature eg. gaussian blur, noise, and jpeg artifacts. \ourmethod{} demonstrates its efficacy on realistic datasets of several physically unrelated degradations which are heavily spatially-varying. In such settings, DuRN and OWAN are quite inferior to our model, as shown in our experiments.

%% file: paper_parts/method2_cameraready.tex
\begin{figure}[!t]
    \centering
    \includegraphics[width = 0.5\textwidth, height=2.4in]{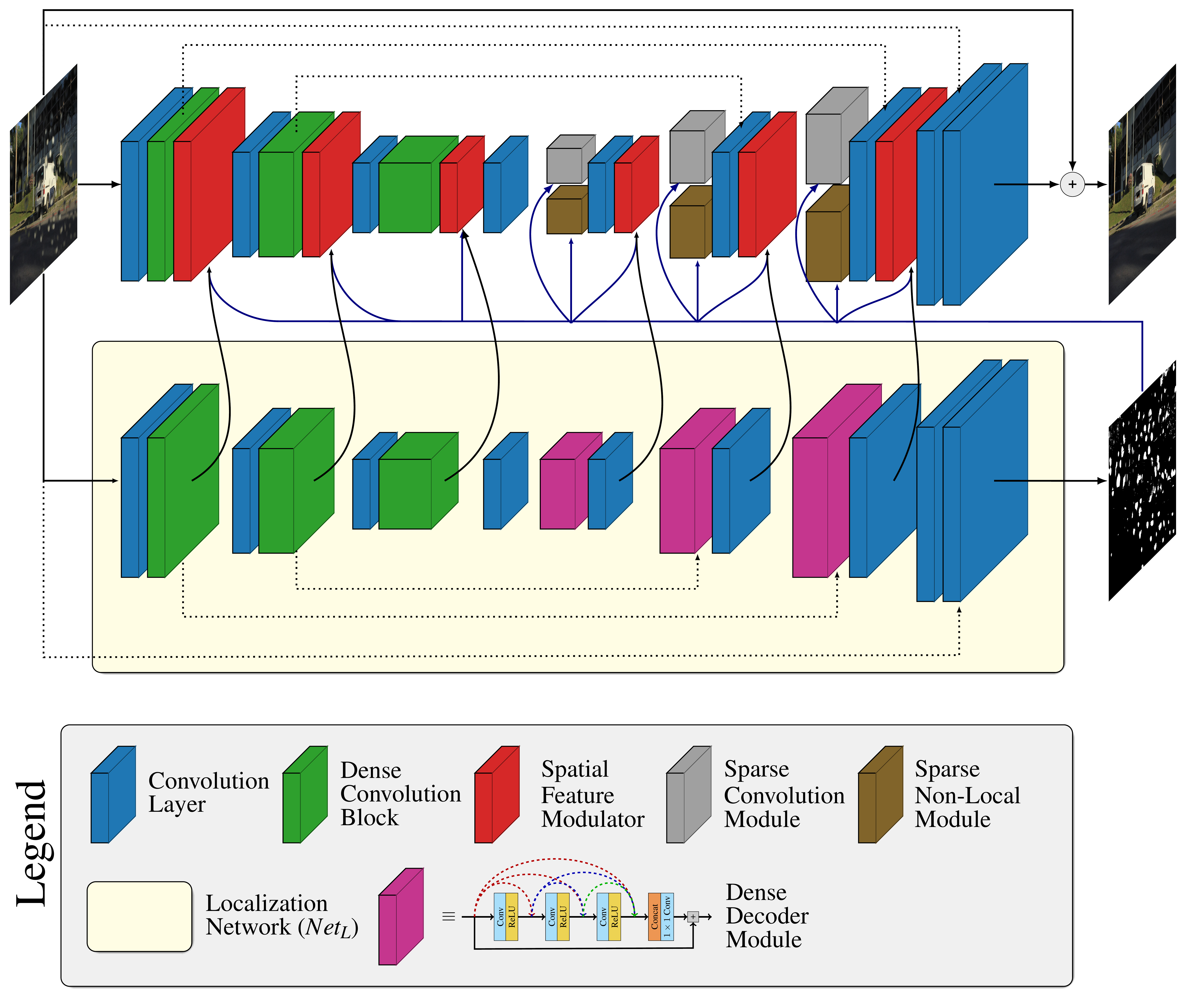}
    \caption{Proposed \ourmethod{} and its components. $Net_R$ is shown at the top and $Net_L$ is shown at the bottom. Connection between the two networks (for SFM) are shown using black arrows.\label{fig:main}}
\end{figure}

An image restoration model needs to solve two equally important tasks: (1) locating the areas to restore in an image, and (2) applying the right filtering mechanism to the corresponding regions. While $Net_L$ addresses the former, we realize the latter through a spatially-guided restoration network $Net_R$. A schematic of \ourmethod{} is shown in Fig. \ref{fig:main}. The knowledge from intermediate features of pre-trained $Net_L$ improves $Net_R$'s training, while the mask itself lends adaptiveness to the restoration process. To realize the twin goals of restoration and reconstruction, distortion-guided filtering of the extracted features in $Net_R$ is enabled through SFM (Spatial Feature Modulator), SC (Sparse Convolution), and SNL (Sparse Non Local) modules.

\subsection{Distortion Localization Network ($Net_L$)}

To maximize the generalizability of our approach, we adopt the U-Net topology \cite{ronneberger2015u} as our CNN backbone (both for localization and restoration networks). Different versions of this are known to be effective for several restoration tasks such as image deblurring \cite{tao2018scale}, denoising \cite{mao2016image}, and general image-to-image translation \cite{isola2017image}. We build a densely connected encoder-decoder structure whose detailed layer-wise description is given in the supplementary. This design delivers competitive performance across all tasks considered and hence acts as a backbone for our $Net_R$ (see Sec. \ref{ablation2}). $Net_L$ is a lightweight version of $Net_R$ (with similar structure) since the binary classification (localization) task is simpler than the intensity regression (restoration) task. 
 
Given a degraded image, $Net_L$ produces a single channel mask and is trained using binary cross-entropy loss to match the GT binary mask. For datasets with no ground truth mask, we use the absolute difference between degraded image and clean image, and threshold it to obtain a binary mask, classifying pixels into degraded (value 1) or clean (value 0). Empirically, we observed that $Net_R$'s performance improves when $Net_L$ is trained to predict only the pixels with severe distortions (as opposed to detecting even minute intensity changes). Note that the distortion-map directly correlates with the difficulty of restoration, and it may differ from the physically occuring degradation-distribution. For instance, when physical rain-steaks are equally distributed throughout the image, the distortion-map would contain more non-zero values in the urban textured regions than the sky regions (since white rain-streaks do not significantly alter the bright intensities in sky). 

\subsection{\textbf{Spatially-Guided Restoration Network ($Net_R$)}}

As depicted in Fig. \ref{fig:main}, $Net_R$ extracts a features pyramid from input degraded image using a cascade of densely connected layers \cite{huang2017densely}. These features are fed to decoder that generates the restored image. Although correction of small intensities changes can be learnt by simple convolutional layers (basic building block for all prior works), they struggle for spatially-distributed heavy degradations. For such regions, localization based guidance improves restoration quality. We propose 3 modules to employ trained $Net_L$ to convey localization knowledge to $Net_R$.

Since image generation process requires decoder to learn both reconstruction and restoration, each level of the decoder contains an SC and an SNL module. Note that we refrain from using SC or SNL in the encoder layers of $Net_R$ since that would completely discard the degraded image intensities (which contain partially-useful information). We employ SFM at multiple levels to perform distortion-guided feature normalization to complement SC and SNL modules.

\subsubsection{\textbf{Spatial Feature Modulator (SFM)}}
\label{sec:knowledge_transfer}

SFM fuses the features of $Net_R$ with intermediate features from layers of the pretrained $Net_L$ in an additive manner. We observe that with such feature guidance, early layers of $Net_R$ extract more distortion-aware features that correlate strongly to the degradation-variation within the input image. Since both the networks share a similar encoder-decoder structure, the inputs of all strided convolution layers are fused using SFM, as shown in Fig. \ref{fig:main}.

In CNNs, feature normalization is known to be important and complementary to feature extraction. The role of SFM is to perform distortion-guided spatial-varying feature normalization. This complements the distortion-guided feature extraction process using local (SC) and global (SNL) context. SFM module performs adaptive shifting of the feature statistics at degraded locations, which aids the restoration process. Studies \cite{johnson2016perceptual} show that feature mean relates to global semantic information while  variance  is correlated  to local texture. Inspired from this, our SFM modulates features at degraded locations to match the feature statistics (mean and variance) of clean regions.

Given the fused features $F$ and the predicted mask $\mathcal{M}$, we calculate the modulated features $F^{S}$ as
\begin{multline}\label{eq:norm1}
F^{S} = \sigma(F,(1-\mathcal{M}))  \left(\frac{F\odot \mathcal{M} - \mu(F,\mathcal{M})}{\sigma(F,\mathcal{M})} \right) \\  + \mu(F,(1-\mathcal{M}))
\end{multline}
The mean operator is $\mu(Q,\mathcal{M}) = \frac{1}{\sum_{p} \mathcal{M}_p} \sum_{p} Q_{p}\odot \mathcal{M}_p$ and the standard deviation is $\sigma(Q,\mathcal{M}) = \sqrt{\frac{1}{\sum_{p} \mathcal{M}_p} \sum_{p} (Q^2_{p}\odot \mathcal{M}_p- \mu(Q,\mathcal{M})) + \epsilon}$, where subscript $p$ represents 2D pixel location and $\odot$ represents element-wise product. Since modulation of features is desired only at the degraded locations, the final feature output of SFM is $F^S \odot \mathcal{M}+ F \odot (1-\mathcal{M})$.

\subsubsection{\textbf{Mask-guided Sparse Convolution (SC)}}

As discussed earlier, filters of general convolution layers are spatially-invariant and hence are forced to learn the restoration and reconstruction tasks jointly, which impedes the training process and reduces model's performance. \ourmethod{} harnesses the efficacy of mask-guided sparse convolution that facilitates selective restoration of highly degraded regions, and simplifies the learning process. SC (shown in Fig. \ref{fig:main}) contains a densely connected set of $6$ guided sparse convolution layers followed by a $1\times1$ convolution to reduce the number of channels. Each unit in SC takes the input feature map, $F$ and the predicted mask $\mathcal{M}$. Pixels masked as $1$ in $\mathcal{M}$ are sampled, and passed through a convolution operation, resulting in a sparse feature map $F^S$ as 
\vspace{-0.1em}
\begin{equation}
F_p^S =\left\{
\begin{array}{rcl}
0       &      & {\mathcal{M}_p = 0}\\
\sum_{p' \in R_k} K_{p'} F_{p+p'}       &      & {\mathcal{M}_{p+p'},\mathcal{M}_p = 1}
\end{array} \right.
\end{equation}
where $R_k$ indicates the support region of kernel offsets with kernel size $k$ (e.g., for a $3\times3$ convolution, $R_k=\{(-1,-1), (-1, 0),...,(1,1)\}$ and $k=3$), and $K\in\mathbb{R}^{C_{in} \times C_{out} \times k \times k}$ denotes convolution weights. Although SC is quite effective for the spatially-varying task at hand, its receptive field is limited to only degraded pixels. We next describe our SNL module which extracts features using global context-aggregation (with distortion-guidance) and complements the role of SC.

\begin{figure}[!t]
    \centering
    \includegraphics[trim={0 0cm 0 0},clip,width = 0.49\textwidth]{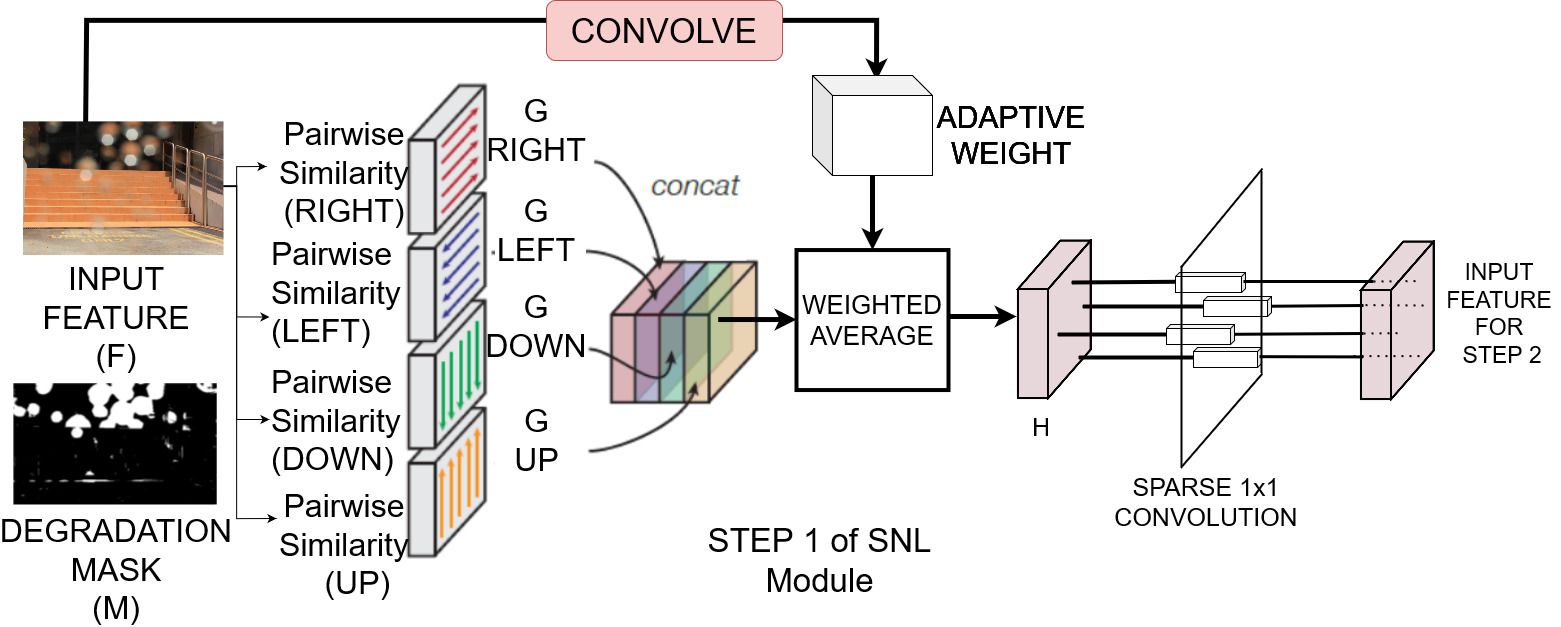}\\
    \caption{Region-Guided Sparse Non-Local (SNL) Module for degradation-guided context aggregation. Sparse 1x1 convolution connects the two structurally-identical sparse-attention steps.}
    \label{fig:rnn}
    \vspace{-0.5cm}
\end{figure}

\subsubsection{\textbf{Region-Guided Sparse Non-Local Module (SNL)}}
Most computer vision tasks are inherently contextual. Frequently used tools for gathering larger context such as dilated convolution \cite{yu2015multi}, global average or attention pooling \cite{hu2018squeeze}, or multi-scale methods \cite{tao2018scale} etc. can enlarge the receptive field beyond simple convolutional layers. Yet, they are not image-adaptive and still cannot utilize the full feature map effectively. In contrast, a single non-local layer \cite{wang2018non} is capable of extending the receptive field to the maximum $H \times W$ size adaptively for each image and each pixel in an image. We claim that such a property is well-suited for a spatially-varying restoration model, where the heavily corrupted regions benefit from the ability to gather relevant features from the whole image. Effectiveness of adaptive global context aggregation has also been explored for recognition/segmentation tasks in \cite{wang2018non,ramachandran2019stand}.

We hypothesize that within a restoration model, the non-local context aggregation process can benefit immensely from the knowledge of degraded pixel locations. We intuit that propagating heavily degraded information throughout the spatial domain can be counter-productive. Ideally, an adaptive module should learn to completely ignore irrelevant features, but recent vision models (e.g. image captioning \cite{huang2019attention}) have shown that this behavior is not practically achieved. They resort to perform additional filtering to remove unnecessary information. 

In contrast, the proposed SNL module leverages the distortion mask to control the scope of non-local context aggregation. While restoring degraded pixels, it assigns dynamically estimated non-zero weights to features from only not/less degraded pixel locations, delivering superior performance as it dampens the influence of heavily degraded/corrupted information. Moreover, SNL leaves the features of clean regions unaltered as this operation is performed only on degraded pixel locations.

As illustrated in Fig. \ref{fig:rnn}, SNL comprises of an efficient two-step aggregation approach, with each step comprising of horizontal-vertical scanning of the feature matrix in four fixed directions: left-to-right, top-to-bottom, and vice-versa. Having two steps is important in harvesting full-image contextual information from all pixels. While directional scanning of CNN features has been explored in literature~\cite{visin2016reseg,liu2016learning}, \ourmethod{} introduces a region-guided and sparse non-local module.

We elaborate on the feature aggregation process within first step of the SNL module for horizontal direction (it can be similarly derived for other directions as well). We denote the value at a particular location $(i,j)$ in the feature map $\mathbf{F} \in \mathbb{R}^{C \times H \times W}$ as $\mathbf{f}_{i,j} \in \mathbb{R}^C$. To model its relationship with all other valid locations (ensuring $\mathcal{M}_{i,j}=0$) on the right $\mathbf{F}_{i,j}^{right} \in \mathbb{R}^{C \times (W-i)}$, we calculate a pairwise relationship matrix $\mathbf{o}^{right} \in \mathbb{R}^{W-i}$ using softmax as
\begin{equation}
\label{eqn:pairwise_relationship}
    \mathbf{o}^{right} = \text{softmax}(\mathbf{f}_{i,j} \odot \mathbf{F}_{i,j}^{right}) 
\end{equation}

This matrix is then used to weigh the contribution of the features towards the right ($\mathbf{F}_{i,j}^{right}$) as
\begin{equation}
\label{eqn:pairwise_relationship2}
    \mathbf{g}_{i,j}^{right} =  \mathbf{F}_{i,j}^{right} \odot \mathbf{o} 
\end{equation}
where $\mathbf{g}_{i,j}^{right} \in \mathbb{R}^C$. Note that locations with $\mathcal{M}_{i,j}=0$ are skipped during the above operations and the four directions are parallely executed in the CUDA implementation. Finally, the features from four directions are fused using pixel-wise adaptive weights. These weights $\mathbf{E} \in \mathbb{R}^{4 \times H \times W}$ are generated by feeding the feature $\mathbf{F}$ to another convolution layer. The fused features $\mathbf{h}_{i,j}$ are obtained as 
\begin{equation}
    \mathbf{h}_{i,j} = \sum_{k \in \Omega}^{4} e_{i,j}^k \odot \mathbf{g}_{i,j}^{k}
\end{equation}
where $e_{i,j}^k \in \mathbb{R}^1$ is the $(k,i,j)$-th element of $\mathbf{E}$ and $k \in \{left,right,up,down\}$. The entire process is repeated twice (Fig. \ref{fig:rnn}) to allow each pixel to gather global context. 

\noindent \textbf{Sparse $1\times1$ convolution:} To perform feature-refinement between two steps of the SNL module, we introduce a sparse $1\times1$ convolution. As shown in the subfigure Fig. \ref{fig:rnn}, on the feature locations of interest specified by the binary mask, a point-wise feature representation is extracted. A fully connected layer then accepts and refines the entire stack of these point-wise features. This replaces the 2D convolution on $H \times W$ spatial grid with point-wise 1D convolution on the selected points and facilitates sparse processing.

\begin{table*}[h]
\begin{center}
\caption{\small Image deraining results. Best and second best scores are \textbf{highlighted} and \underline{underlined}. For each method, relative MSE reduction achieved by \ourmethod{} is reported in parenthesis (see Section~\ref{Implementation} for calculation). \ourmethod{} achieves $\sim$$22\%$ improvement over MSPFN~\cite{mspfn2020}.}
\label{table:deraining}
\setlength{\tabcolsep}{5.9pt}
\scalebox{0.74}{
\begin{tabular}{l c c c c c c c c c c || c c}
\toprule[0.15em]
  & \multicolumn{2}{c}{Test100~\cite{zhang2019image}}&\multicolumn{2}{c}{Rain100H~\cite{yang2017deep}}&\multicolumn{2}{c}{Rain100L~\cite{yang2017deep}}&\multicolumn{2}{c}{Test2800~\cite{fu2017removing}}&\multicolumn{2}{c||}{Test1200~\cite{zhang2018density}}&\multicolumn{2}{c}{Average}\\
 Methods & PSNR~$\textcolor{black}{\uparrow}$ & SSIM~$\textcolor{black}{\uparrow}$ & PSNR~$\textcolor{black}{\uparrow}$ & SSIM~$\textcolor{black}{\uparrow}$ & PSNR~$\textcolor{black}{\uparrow}$ & SSIM~$\textcolor{black}{\uparrow}$ & PSNR~$\textcolor{black}{\uparrow}$ & SSIM~$\textcolor{black}{\uparrow}$ & PSNR~$\textcolor{black}{\uparrow}$ & SSIM~$\textcolor{black}{\uparrow}$ & PSNR~$\textcolor{black}{\uparrow}$ & SSIM~$\textcolor{black}{\uparrow}$\\
\midrule
DerainNet~\cite{fu2017clearing} & 22.77  & 0.810  & 14.92  & 0.592  & 27.03  & 0.884  & 24.31  & 0.861  & 23.38  & 0.835  & 22.48 \colorbox{gray!20}{(69.3\%)}  & 0.796  \colorbox{gray!20}{(61.3\%)}  \\
SEMI~\cite{wei2019semi} & 22.35&0.788& 16.56&0.486& 25.03&0.842& 24.43&0.782& 26.05&0.822& 22.88 \colorbox{gray!20}{(67.8\%)}&0.744 \colorbox{gray!20}{(69.1\%)}\\
DIDMDN~\cite{zhang2018density} & 22.56&0.818& 17.35&0.524& 25.23&0.741& 28.13&0.867& 29.65&0.901& 24.58 \colorbox{gray!20}{(60.9\%)}&0.770 \colorbox{gray!20}{(65.7\%)} \\
UMRL~\cite{yasarla2019uncertainty} & 24.41&0.829& 26.01&0.832& 29.18&0.923& 29.97&0.905& 30.55&0.910& 28.02 \colorbox{gray!20}{(41.9\%)}&0.880 \colorbox{gray!20}{(34.2\%)} \\
RESCAN~\cite{li2018recurrent} & 25.00&0.835& 26.36&0.786& 29.80&0.881& 31.29&0.904& 30.51&0.882& 28.59 \colorbox{gray!20}{(37.9\%)} & 0.857 \colorbox{gray!20}{(44.8\%)} \\
PreNet~\cite{ren2019progressive} & 24.81&0.851& 26.77&0.858& \underline{32.44} & \underline{0.950} & 31.75&0.916& 31.36&0.911& 29.42 \colorbox{gray!20}{(31.7\%)} &0.897 \colorbox{gray!20}{(23.3\%)}\\
MSPFN~\cite{mspfn2020}  & \underline{27.50} & \underline{0.876} & \underline{28.66} & \underline{0.860} & 32.40&0.933& \underline{32.82} & \underline{0.930} & 32.39 & \underline{0.916} & \underline{30.75} \colorbox{gray!20}{(21.9\%)} & \underline{0.903} \colorbox{gray!20}{(18.6\%)} \\
\textbf{\ourmethod{} (Ours)}  & \textbf{30.35} & \textbf{0.909} & \textbf{30.95} & \textbf{0.892} & \textbf{36.93} & \textbf{0.969} & \textbf{33.34} & \textbf{0.936} & \textbf{33.04} & \textbf{0.922} & \textbf{32.91} \colorbox{gray!20}{(0.0\%)}& \textbf{0.926} \colorbox{gray!20}{(0.0\%)}\\

\bottomrule[0.1em]
\end{tabular}}
\end{center}\vspace{-1.3em}
\end{table*}

\begin{table*}[h!]
\centering
\caption{Quantitative comparisons of models trained and tested on the SPANet~\cite{wang2019spatial} and the Rain100H~\cite{yang2017deep} benchmarks.\label{table:spanet_comparisons}}
\resizebox{\textwidth}{!}{%
\begin{tabular}{@{}lllllllllllllll@{}}
\toprule
Dataset & Metric  & DSC   & GMM  & Clear  & DDN    & RESCAN &  PReNet  &  {SPANet}  & JORDER$_E$ & RCDNet$_1$ & RCDNet &  \textbf{\ourmethod{}}\\ \midrule
\multirow{2}{*}{SpaNet~\cite{wang2019spatial}}&PSNR  & 34.95 & 34.30 &  34.39  & 36.16 & 38.11 & 40.16  & {40.24} & 40.78 & 40.99 & 41.47 & {\bf 44.10}\\
    & SSIM & 0.9416 & 0.9428  & 0.9509 & 0.9463  & 0.9707 &  0.9816 & 0.9811 & { **} & 0.9816 & 0.9834 & {\bf 0.9872}\\
\multirow{2}{*}{Rain100H~\cite{yang2017deep}}& PSNR  & 13.77 & 15.23 &  15.33  & 22.85 & 29.62 & 30.11  & {25.11} & 30.50 & 30.91 & 31.28 & {\bf 31.69} \\ 
& SSIM & 0.3199 & 0.4498  & 0.7421 & 0.7250  & 0.8720 &  0.9053 & 0.8332 & { 0.8967} & 0.9037 & 0.9093 & {\bf 0.9201} \\ 
  \bottomrule
\end{tabular}
}
 \vspace{-0.4cm}
\end{table*}

\begin{figure*}[t]
\setlength{\tabcolsep}{2pt}
\footnotesize
\begin{tabular}{ccccccccc}
        
        \includegraphics[height=1.2cm,width=0.1\textwidth]{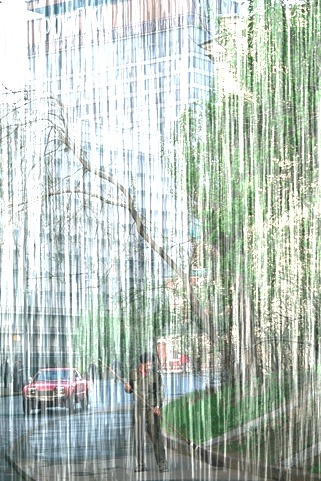} &
        \includegraphics[bb=30 280 150 365,clip=True,width=.1\textwidth]{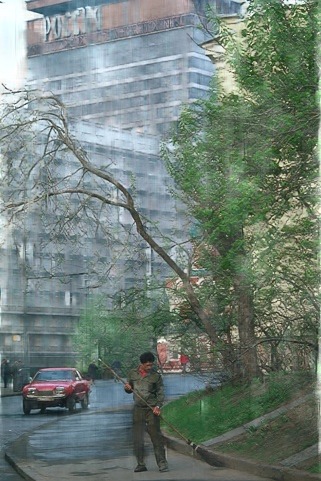} &        
		\includegraphics[bb=30 280 150 365,clip=True,width=.1\textwidth]{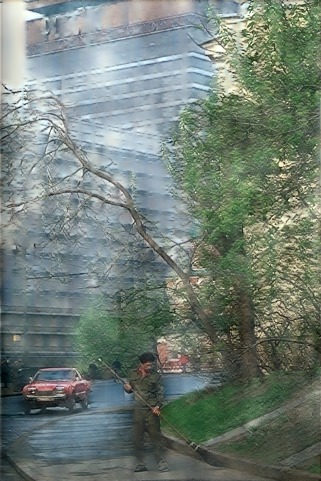} &
		\includegraphics[bb=30 280 150 365,clip=True,width=.1\textwidth]{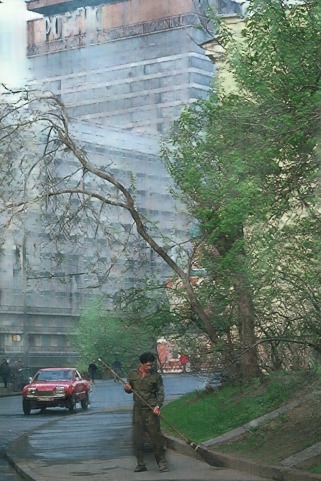} &
		\includegraphics[bb=30 280 150 365,clip=True,width=.1\textwidth]{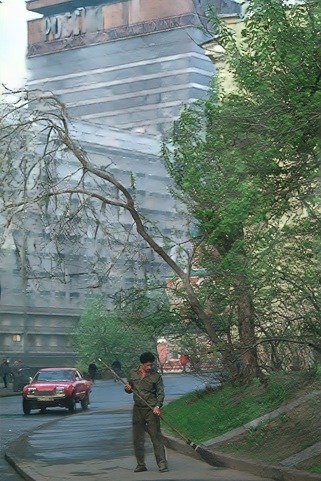} &
		\includegraphics[bb=30 280 150 365,clip=True,width=.1\textwidth]{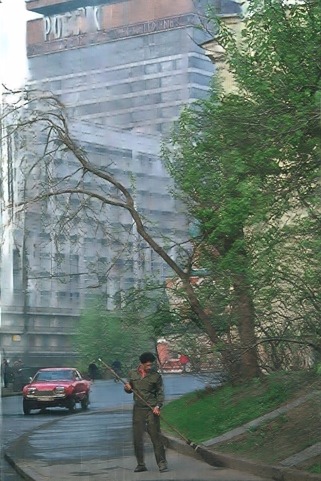} &		
		\includegraphics[bb=30 280 150 365,clip=True,width=.1\textwidth]{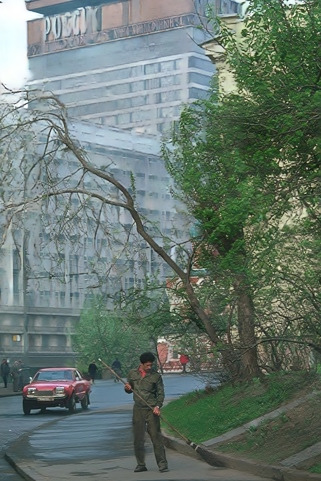} &
		\includegraphics[bb=30 280 150 365,clip=True,width=.1\textwidth]{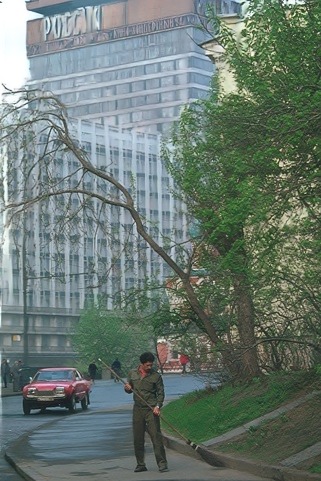} &
		\includegraphics[bb=30 280 150 365,clip=True,width=.1\textwidth]{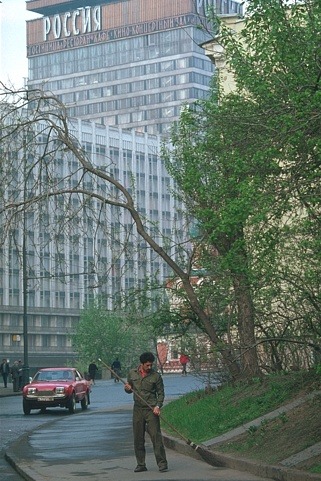} \\     
        
        \includegraphics[height=1.1cm,width=.1\textwidth]{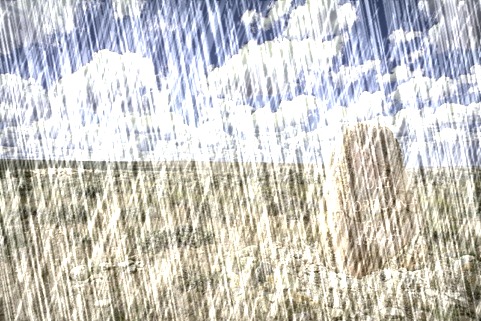} &
        \includegraphics[bb=300 70 430 150,clip=True,width=.1\textwidth]{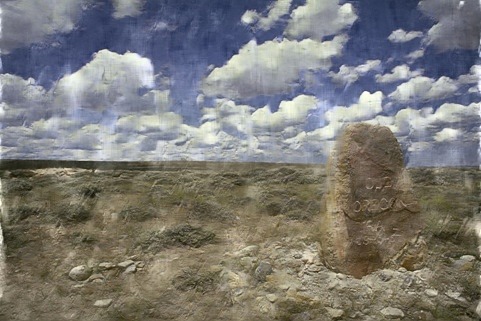} &
        \includegraphics[bb=300 70 430 150,clip=True,width=.1\textwidth]{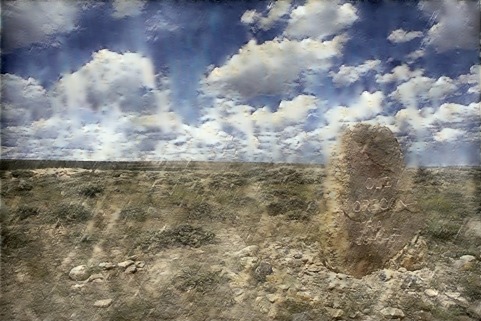} &        
        \includegraphics[bb=300 70 430 150,clip=True,width=.1\textwidth]{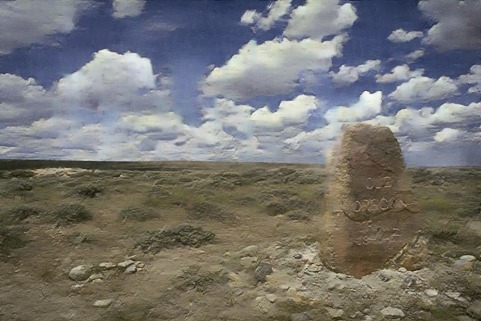} &
        \includegraphics[bb=300 70 430 150,clip=True,width=.1\textwidth]{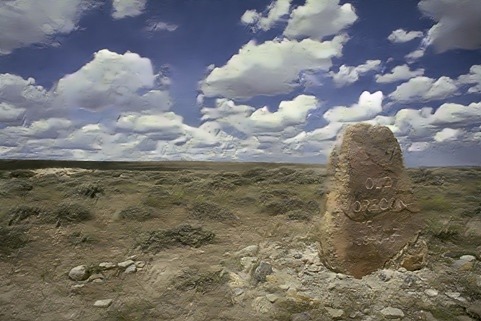} &
        \includegraphics[bb=300 70 430 150,clip=True,width=.1\textwidth]{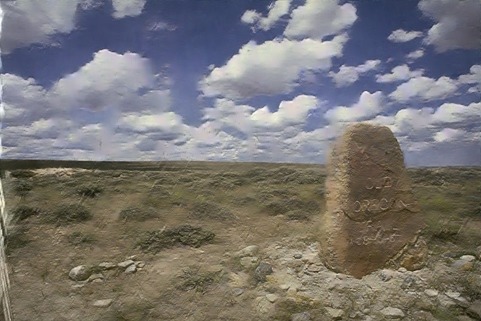} &
        \includegraphics[bb=300 70 430 150,clip=True,width=.1\textwidth]{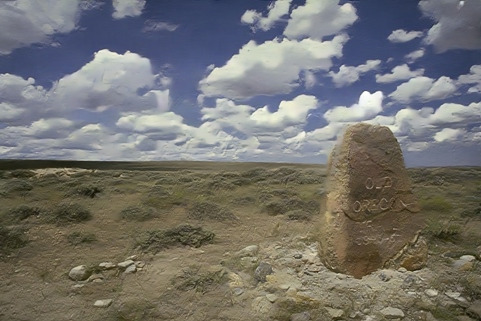} &
        \includegraphics[bb=300 70 430 150,clip=True,width=.1\textwidth]{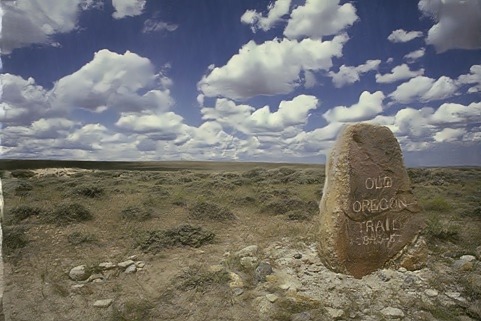} &
        \includegraphics[bb=300 70 430 150,clip=True,width=.1\textwidth]{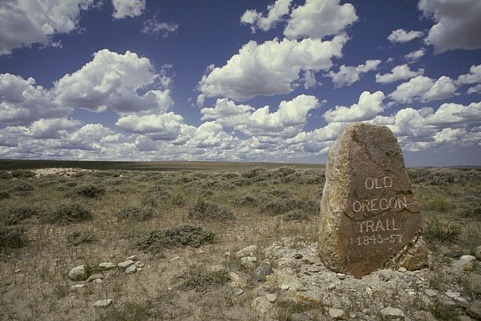} \\
        
\footnotesize{Input} &  JORDER \cite{yang2017deep} &  Fu et al. \cite{fu2017removing} &  RESCAN \cite{li2018recurrent} &  PReNet \cite{ren2019progressive} &  SPANet \cite{wang2019spatial} &  RDCNet \cite{wang2020model}  &  \ourmethod{} & GT \\
    \end{tabular}
    
    \caption{Qualitative comparison of zoomed-in results on synthetic rainy images from the Rain100H test-set.}
    \label{fig:results_heavy}
\end{figure*}


\begin{figure*}
\footnotesize
\centering
\setlength{\tabcolsep}{0pt}
\begin{tabular}{ccccc|cccccc}
         \includegraphics[width=.095\textwidth]{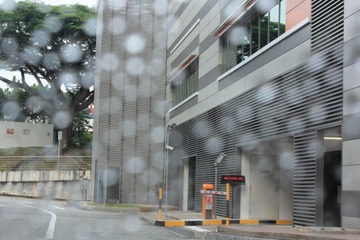} &
         \includegraphics[width=.095\textwidth]{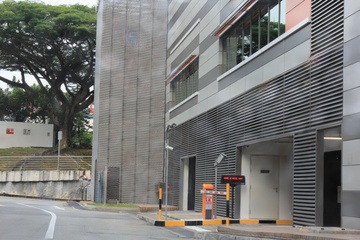} &
        \includegraphics[width=.095\textwidth]{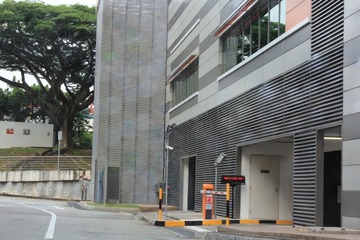} &
        \includegraphics[width=.095\textwidth]{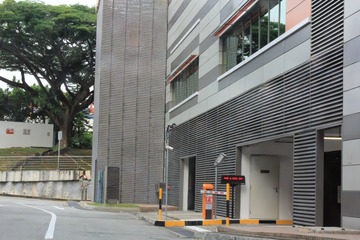} &
        \includegraphics[width=.095\textwidth]{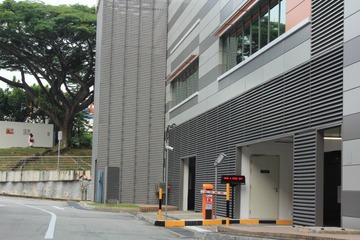} \hspace{0.1em}
 &
\hspace{0.1em}
         \includegraphics[width=.095\textwidth]{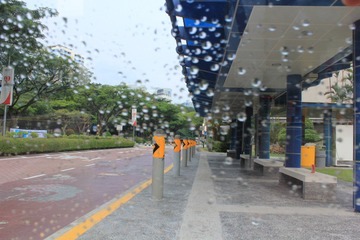} &
         \includegraphics[width=.095\textwidth]{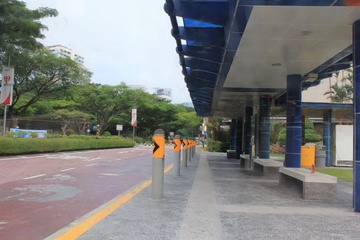} &
        \includegraphics[width=.095\textwidth]{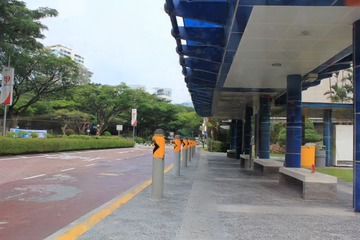} &
        \includegraphics[width=.095\textwidth]{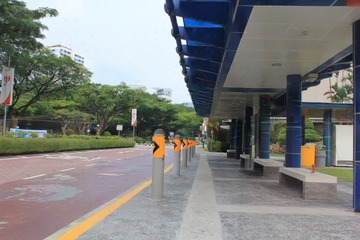} &
        \includegraphics[width=.095\textwidth]{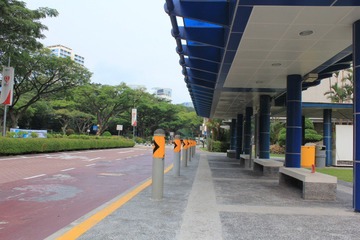} \\
        
         \includegraphics[width=.095\textwidth]{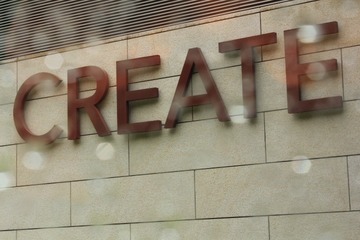} &
         \includegraphics[width=.095\textwidth]{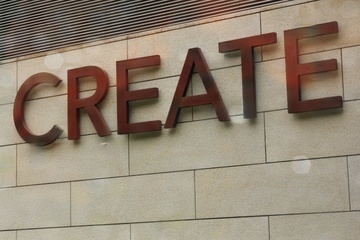} &
        \includegraphics[width=.095\textwidth]{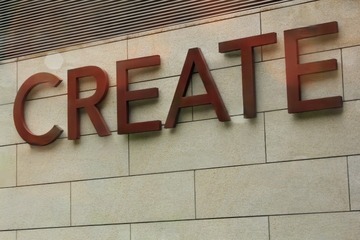} &
        \includegraphics[width=.095\textwidth]{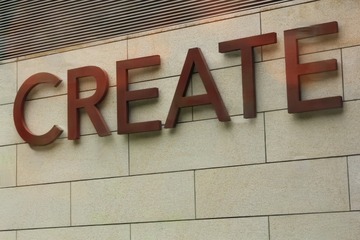} &
        \includegraphics[width=.095\textwidth]{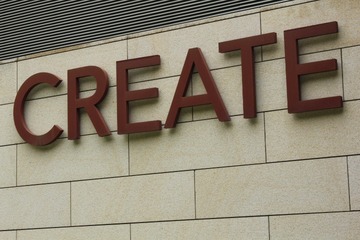} \hspace{0.1em}
 &
\hspace{0.1em}
         \includegraphics[width=.095\textwidth]{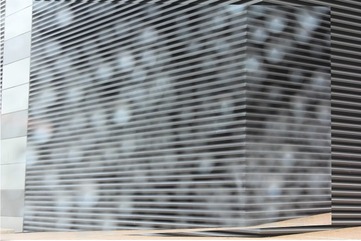} &
         \includegraphics[width=.095\textwidth]{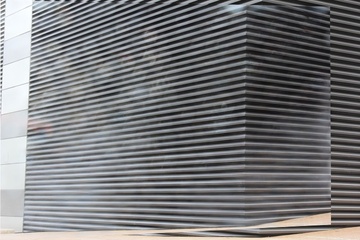} &
        \includegraphics[width=.095\textwidth]{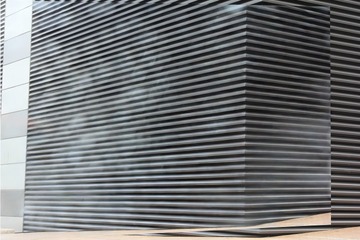} &
        \includegraphics[width=.095\textwidth]{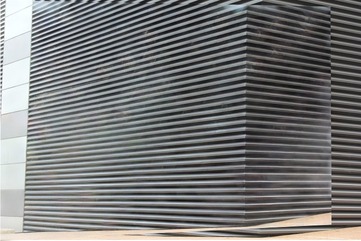} &
        \includegraphics[width=.095\textwidth]{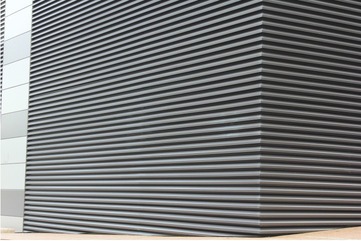} \\
             Input &  AGAN \cite{qian2018attentive} &  DuRN \cite{DuRN_cvpr19} &  \ourmethod{}  & GT &  Input &  AGAN \cite{qian2018attentive} &  DuRN \cite{DuRN_cvpr19} &  \ourmethod{}  & GT \\
        
    \end{tabular}
    \caption{Qualitative comparisons of results on images from the AGAN testset~\cite{qian2018attentive}.\label{fig:results_raindrop}}
 \vspace{-0.4cm}
\end{figure*}

%% file: paper_parts/rainstreak.tex
\noindent \textbf{Rain-streak Removal:}  Following prior art~\cite{mspfn2020}, we perform quantitative evaluations (PSNR/SSIM scores) on the Y channel (in YCbCr color space). 
Table~\ref{table:deraining} reports the results across all five datasets where \ourmethod{} consistently achieves significant gains over the baselines.
Compared to the recent algorithm MSPFN~\cite{mspfn2020}, we obtain a performance gain of $2.16$~dB (averaged across all datasets). 
Next, for fair comparison with RCDNet~\cite{wang2020model}, we evaluate \ourmethod{} in their setting (in Table \ref{table:spanet_comparisons}) by training and testing on the challenging Rain100H~\cite{yang2017deep} and SPANet~\cite{wang2019spatial} (captured in real-world rainy scenes) datasets.  While the improvement is modest (0.41 dB) on very heavy rain (Rain100H), it is as large as ${3}$~dB on datasets with low rain density, eg. SPANet and Rain100L (since in this case, we selectively process very few pixels without affecting clean pixels), highlighting the advantage of our distortion-adaptive restoration.

Fig. ~\ref{fig:results_heavy} presents qualitative comparisons on challenging images from Rain100H dataset. Our results exhibit significantly higher visual quality than existing methods which fail to recover background textures (1$^{st}$ row) and introduce artifacts (2$^{nd}$ row). \ourmethod{} is robust to changes in scenes and rain densities as it effectively removes rain streaks of different orientations and magnitudes, and generates images that are visually pleasing and faithful to the ground-truth.

%% file: paper_parts/raindrop.tex
\begin{table}[t]
	\centering
	\setlength{\tabcolsep}{3pt}
\caption{Raindrop removal results on testset from Qian et al. \cite{qian2018attentive}.}	

\resizebox{1\linewidth}{!}{%

	\begin{tabular}{lllllll}
		\toprule
		Method & Eigen \cite{eigen2013restoring} &  Pix2pix \cite{isola2017image}     & AGAN\cite{qian2018attentive} &  DuRN\cite{DuRN_cvpr19} & Quan\cite{quan2019deep} & \ourmethod{}  \\
		\hline
		PSNR   & 28.59   &  30.59   &  31.51   &  31.24   & 31.44   &  \bf{32.73}   \\
		SSIM   & 0.6726   &  0.8075   &  0.9213   &  0.9259   & 0.9263   &  \bf{0.9410} \\
		\bottomrule
	\end{tabular}
	}
	\label{table:raindrop_dataset}
 \vspace{-0.4cm}
\end{table}

\vspace{3pt}
\noindent \textbf{Raindrop Removal:} Table~\ref{table:raindrop_dataset} and Fig. \ref{fig:results_raindrop} show qualitative and visual comparisons with recent methods \cite{qian2018attentive,quan2019deep,DuRN_cvpr19}. \ourmethod{} outperforms the baselines by a large margin. Our results are visually closer to GT and perceptually better than those of competing methods which often contain artifacts or color distortions.

%% file: paper_parts/shadow.tex
\begin{figure*}[h]
\footnotesize

  \centering
    \setlength{\tabcolsep}{1pt}
  			\begin{tabular}{cccccccc}
  \includegraphics[width=\widthscalesix\linewidth,height = \widthscalefive\linewidth]{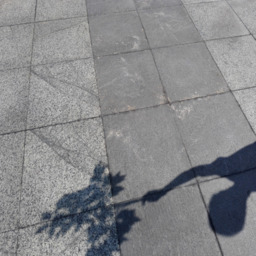}&
  \includegraphics[bb=180 10 256 100,clip=True,width=\widthscalesix\linewidth,height = \widthscalefive\linewidth]{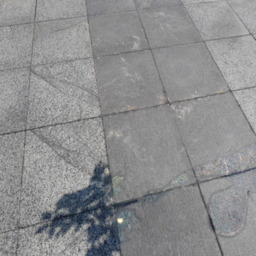}&
  \includegraphics[bb=180 10 256 100,clip=True,width=\widthscalesix\linewidth,height = \widthscalefive\linewidth]{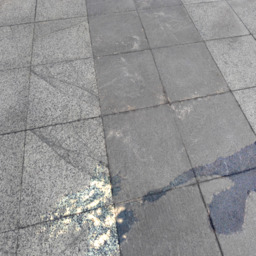}&
  \includegraphics[bb=180 10 256 100,clip=True,width=\widthscalesix\linewidth,height = \widthscalefive\linewidth]{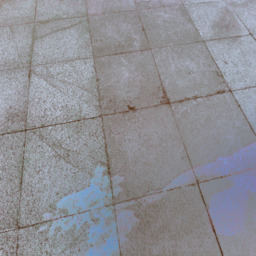}&
  \includegraphics[bb=180 10 256 100,clip=True,width=\widthscalesix\linewidth,height = \widthscalefive\linewidth]{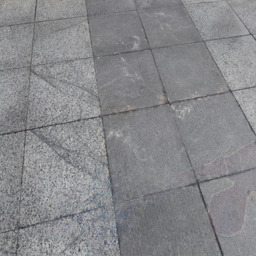}&
  \includegraphics[bb=180 10 256 100,clip=True,width=\widthscalesix\linewidth,height = \widthscalefive\linewidth]{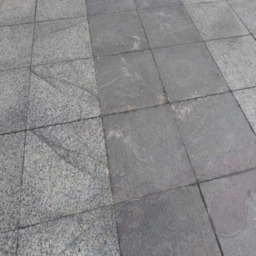}&
  \includegraphics[bb=180 10 256 100,clip=True,width=\widthscalesix\linewidth,height = \widthscalefive\linewidth]{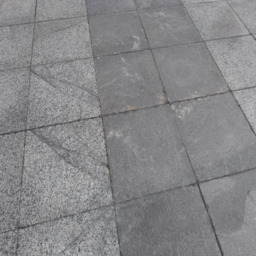}&
  \includegraphics[bb=180 10 256 100,clip=True,width=\widthscalesix\linewidth,height = \widthscalefive\linewidth]{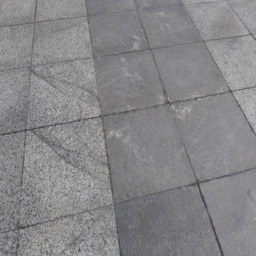}\\
    \includegraphics[width=\widthscalesix\linewidth,height = \widthscalefive\linewidth]{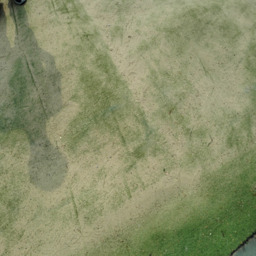}&
  \includegraphics[bb=10 40 106 120,clip=True,width=\widthscalesix\linewidth,height = \widthscalefive\linewidth]{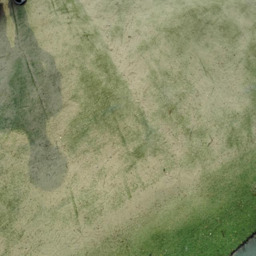}&
  \includegraphics[bb=10 40 106 120,clip=True,width=\widthscalesix\linewidth,height = \widthscalefive\linewidth]{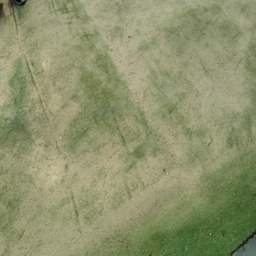}&
  \includegraphics[bb=10 40 106 120,clip=True,width=\widthscalesix\linewidth,height = \widthscalefive\linewidth]{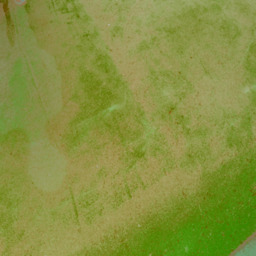}&
  \includegraphics[bb=10 40 106 120,clip=True,width=\widthscalesix\linewidth,height = \widthscalefive\linewidth]{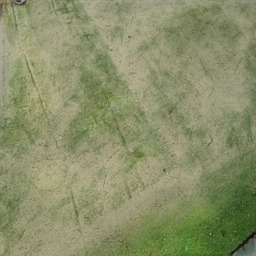}&
  \includegraphics[bb=10 40 106 120,clip=True,width=\widthscalesix\linewidth,height = \widthscalefive\linewidth]{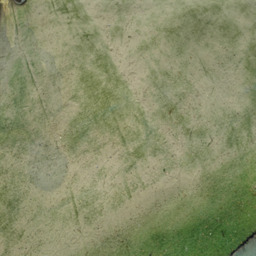}&
  \includegraphics[bb=10 40 106 120,clip=True,width=\widthscalesix\linewidth,height = \widthscalefive\linewidth]{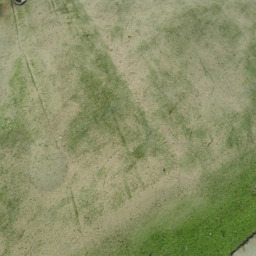}&
  \includegraphics[bb=10 40 106 120,clip=True,width=\widthscalesix\linewidth,height = \widthscalefive\linewidth]{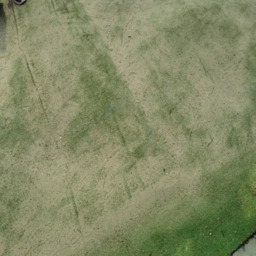}\\
\footnotesize Input &  \cite{guo2012paired}     & \cite{gong2014interactive} & \cite{yang2012shadow} &  \cite{wang2018stacked} & \cite{hu2018direction} & \cite{cun2020towards} & \ourmethod{}  \\
  \end{tabular}
\caption{Comparison of shadow removal results on ISTD Dataset~\cite{wang2018stacked}. Shadow region and boundaries are visible in existing approaches.}
\label{fig:shadow_qual}
\end{figure*}

\noindent \textbf{Shadow Removal:} We evaluate our shadow-removal model against traditional \cite{guo2012paired,gong2014interactive,yang2012shadow} and learning based methods including ST-CGAN~\cite{wang2018stacked}, DSC~\cite{hu2018direction}, DeShadowNet~\cite{qu2017deshadownet}. Following prior art, results are evaluated in Lab color space using RMSE scores calculated over shadow and non-shadow regions. Fig. \ref{fig:shadow_qual} and Table \ref{table:shadow_quant} show that although CNN-based designs are better than hand-crafted methods, most existing approaches produce  shadow boundaries or color inconsistencies. 
However, \ourmethod{} has minimal artifacts in the shadow boundaries, outperforming the baselines both qualitatively and quantitatively.

\begin{table}[t]
	\centering
	\setlength{\tabcolsep}{3pt}
\caption{Shadow removal results on ISTD Dataset~\cite{wang2018stacked}. Subscripts S and NS indicate shadow and non-shadow regions, respectively.}
\resizebox{1\linewidth}{!}{%
	\begin{tabular}{lllllllllrr}
		\toprule
		Metric & Input & \cite{yang2012shadow} &   \cite{guo2012paired}     & \cite{gong2014interactive} &  \cite{wang2018stacked} & \cite{hu2018direction}& \cite{zhang2020ris}& \cite{cun2020towards} & \ourmethod{}  \\
		\midrule
		RMSE$_S$   & 32.12  &  19.82   &  18.95   &  14.98   & 10.33 & 9.48 &  8.99 & 8.14  &  \bf{8.05}   \\
		RMSE$_{NS}$  & 7.19   &  14.83   &  7.46   &  7.29  & 6.93 &6.14 & 6.33 & 6.04   &  \bf{5.47} \\
		RMSE     & 10.97 & 15.63   &  9.30   &  8.53 & 7.47 & 6.67 & 6.95 & 6.37   &  \bf{5.88}   \\
		\bottomrule
	\end{tabular}
	}
	\label{table:shadow_quant}
 \vspace{-0.3cm}
\end{table}

%% file: paper_parts/blur_cameraready.tex
\begin{table}[t]
\begin{center}
\caption{\small Deblurring results. Our method is trained only on the GoPro dataset~\cite{gopro2017} and directly applied to the test images of HIDE~\cite{shen2019human} and RealBlur-J~\cite{rim_2020_realblur} datasets. PSNR\textcolor{red}{$\ddagger$} scores were obtained after training and testing on RealBlur-J dataset.}
\label{table:gopro and hide}
\setlength{\tabcolsep}{2pt}
\scalebox{0.83}{
\begin{tabular}{l c  c || c  c || c c  c }
\toprule[0.15em]
 & \multicolumn{2}{c||}{GoPro~\cite{gopro2017}} & \multicolumn{2}{c||}{HIDE~\cite{shen2019human}} & \multicolumn{3}{c}{RealBlur-J~\cite{rim_2020_realblur}} \\
 Method & PSNR & SSIM & PSNR & SSIM &
 PSNR &
 SSIM &
  PSNR\textcolor{red}{$^\ddagger$}\\
\midrule[0.15em]
Xu \etal \cite{xu2013unnatural}     & 21.00  & 0.741  & - & -  &  27.14   &  0.830  \\
DeblurGAN \cite{deblurgan}          & 28.70  & 0.858  & 24.51  &  0.871 &  27.97  &  0.834  \\
Nah \etal \cite{gopro2017}          & 29.08  & 0.914  & 25.73 & 0.874 &  27.87   &  0.827  \\
Zhang \etal \cite{zhang2018dynamic} & 29.19  & 0.931  & - & - &  27.80   &  0.847  \\
\small{DeblurGAN-v2 \cite{deblurganv2}}    & 29.55  & 0.934  & 26.61  & 0.875 &  \underline{28.70}   &  0.866 & 29.69 \\
SRN~\cite{tao2018scale}             & 30.26  & 0.934  & 28.36  & 0.915  &  28.56   &  \underline{0.867} & \underline{31.38} \\
Shen \etal \cite{shen2019human}     & 30.26 & 0.940 & 28.89  & 0.930 & - & -  \\
Purohit \etal \cite{purohit2019bringing}         & 30.58  & 0.941  & -  & - & - & -  \\
Purohit \etal \cite{purohit2019efficient}         & 30.73  & 0.942  & -  & - & - & -  \\
DBGAN \cite{zhang2020dbgan}         & 31.10  & 0.942  & 28.94  & 0.915 & - & -  \\
MT-RNN \cite{mtrnn2020}             & 31.15 & 0.945 & 29.15 & 0.918 & - & - \\
DMPHN \cite{dmphn2019}              & 31.20  & 0.940  & 29.09  & 0.924 & 28.42   &  0.860  \\
RADN \cite{purohit2020region}              & 31.76  & \underline{0.952}  & 29.68  & 0.927 & -   & -  \\
Suin \etal \cite{Maitreya2020}      & \underline{31.85}  & 0.948  & \underline{29.98}  & \underline{0.930} & - & -  \\
\bottomrule[0.1em]
\textbf{\ourmethod{}} & \textbf{32.06}  & \textbf{0.953}  &	\textbf{30.29} 	&\textbf{0.931} & \textbf{28.81}  & \textbf{0.875} & \textbf{31.82} \\
\bottomrule[0.1em]
\end{tabular}}
\end{center}\vspace{-2em}
\end{table}

\vspace{3pt}
\noindent \textbf{Deblurring:} We validate our distortion-guided approach for general motion deblurring on 3 benchmarks: GoPro~\cite{gopro2017}, HIDE~\cite{shen2019human}, and the real-world blurred images of a recent RealBlur-J~\cite{rim_2020_realblur}. We report the quantitative comparisons with the existing deblurring approaches in Table~\ref{table:gopro and hide}.  
Overall, \ourmethod{} performs favorably against other algorithms. Note that inspite of training only on the GoPro, it outperforms all methods including \cite{shen2019human} on HIDE, without requiring any human bounding box supervision, thereby demonstrating its strong generalization capability.

We evaluate models on RealBlur-J~\cite{rim_2020_realblur} testset under two experimental settings:~1) training on GoPro (to test generalization to real images), and 2) training on RealBlur-J. \ourmethod{} obtains performance gain of $0.39$ dB over the DMPHN model~\cite{dmphn2019} in setting $1$, and $0.44$ dB over existing best method for setting $2$. Our model's effectiveness is owed to the robustness of the distortion-aware approach.

\begin{figure*}[h] \label{fig:visual_deblur}
	\scriptsize
	\centering
		\footnotesize
		\setlength{\tabcolsep}{2pt}
			\begin{tabular}{ccccccccc}
		
				\includegraphics[width=0.12\textwidth, height=0.060\textwidth]{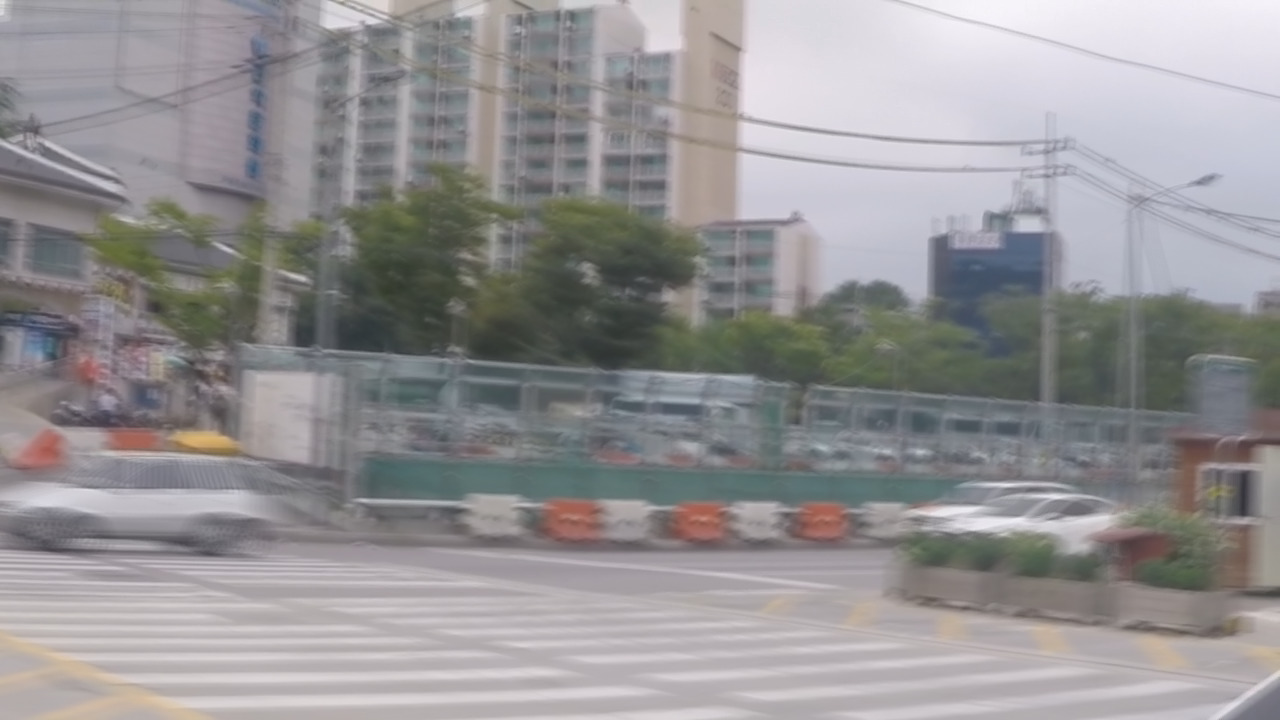}
				&
				\includegraphics[bb=10 150 290 300,clip=True,width=\widthscaleblur \textwidth]{deblurring2/000116_blur} & 
				\includegraphics[bb=10 140 290 290,clip=True,width=\widthscaleblur \textwidth]{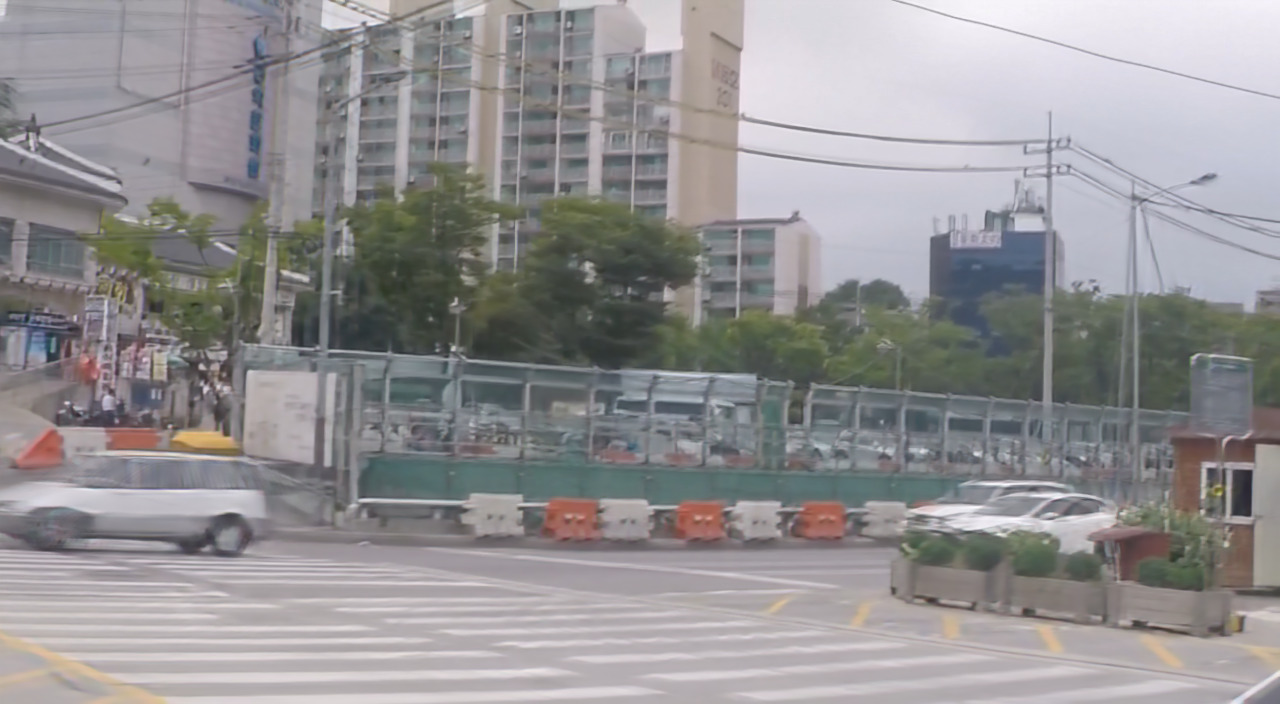} & 
				\includegraphics[bb=10 150 290 300,clip=True,width=\widthscaleblur \textwidth]{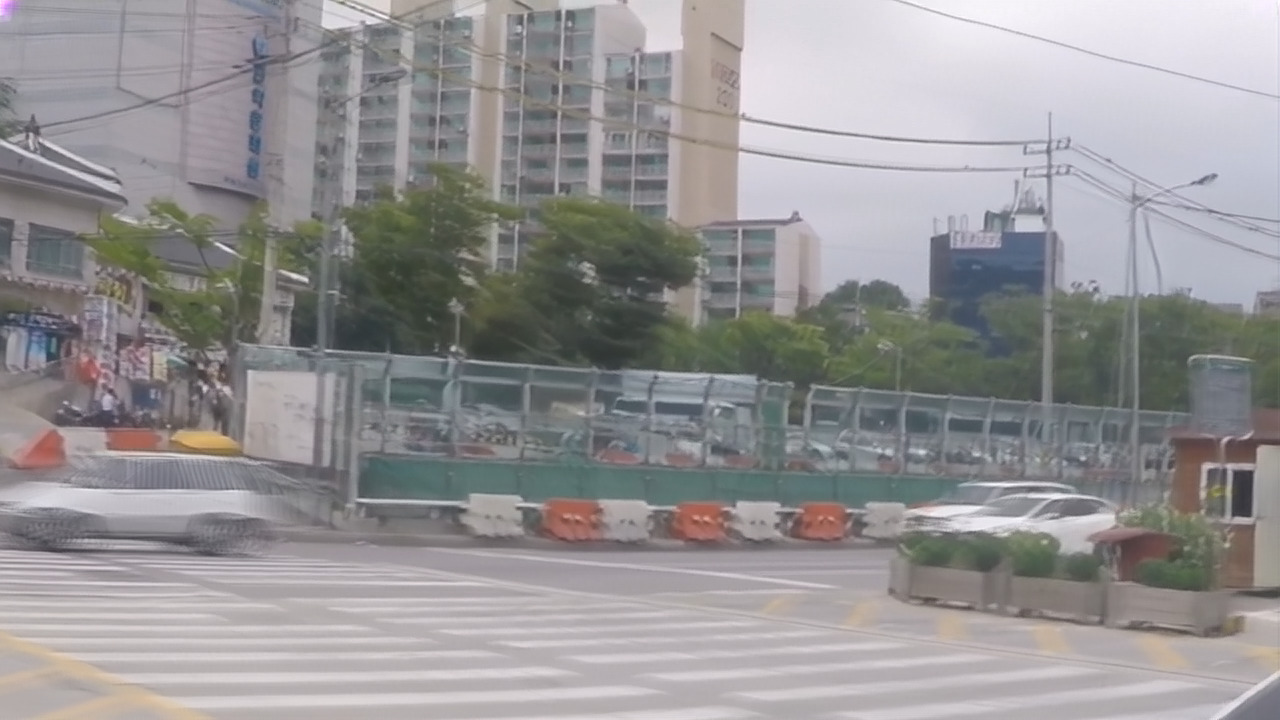} &
				\includegraphics[bb=10 150 290 300,clip=True,width=\widthscaleblur \textwidth]{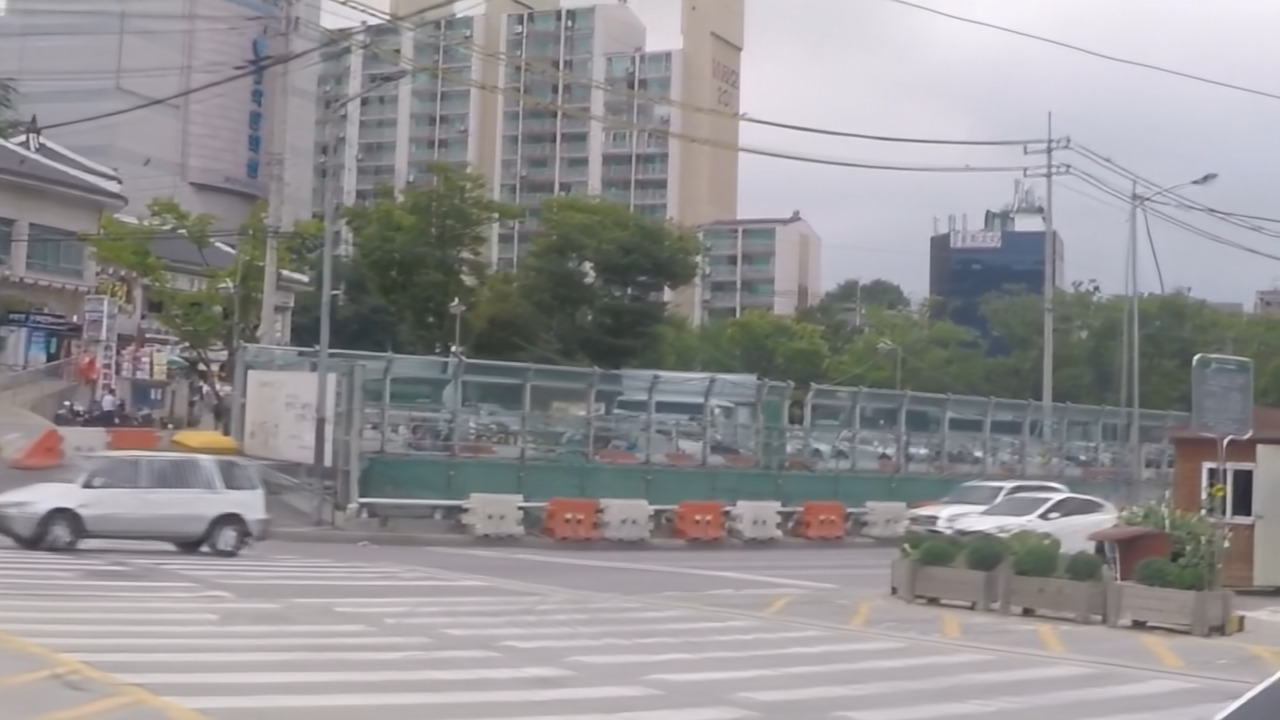} &
				\includegraphics[bb=10 150 290 300,clip=True,width=\widthscaleblur \textwidth]{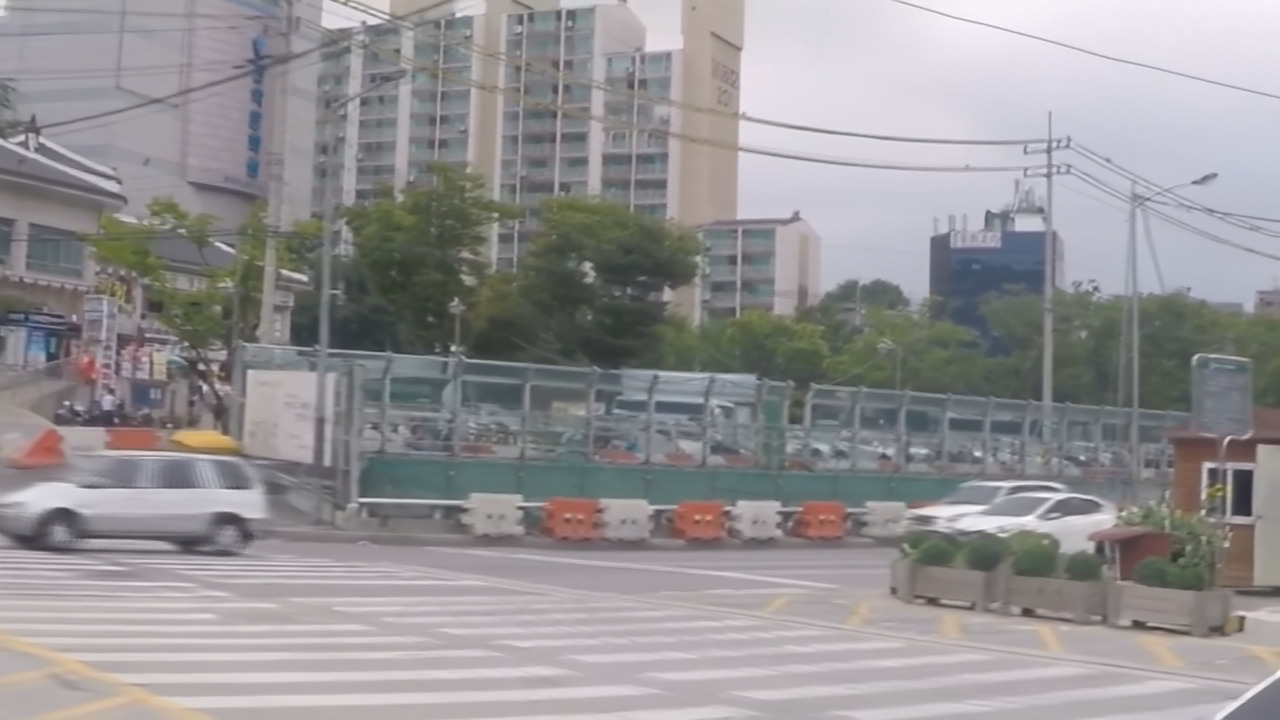} &				 
				\includegraphics[bb=10 150 290 300,clip=True,width=\widthscaleblur \textwidth]{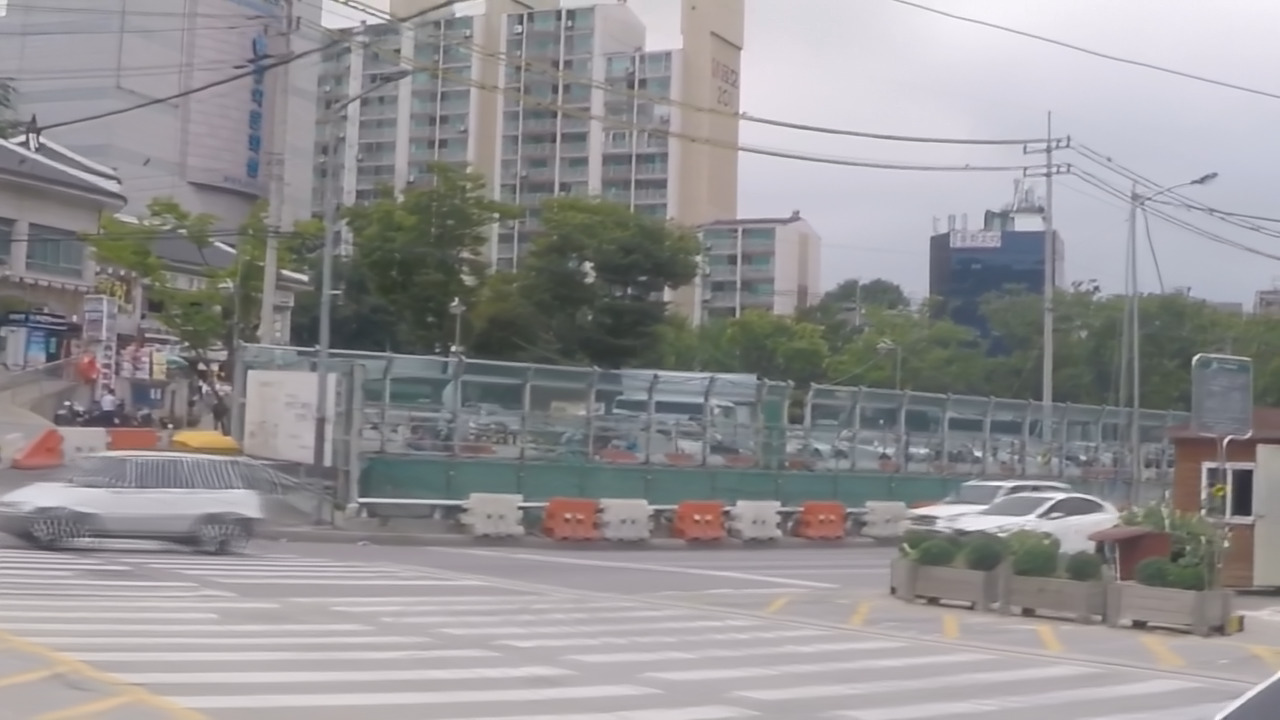} &
				\includegraphics[bb=10 150 290 300,clip=True,width=\widthscaleblur \textwidth]{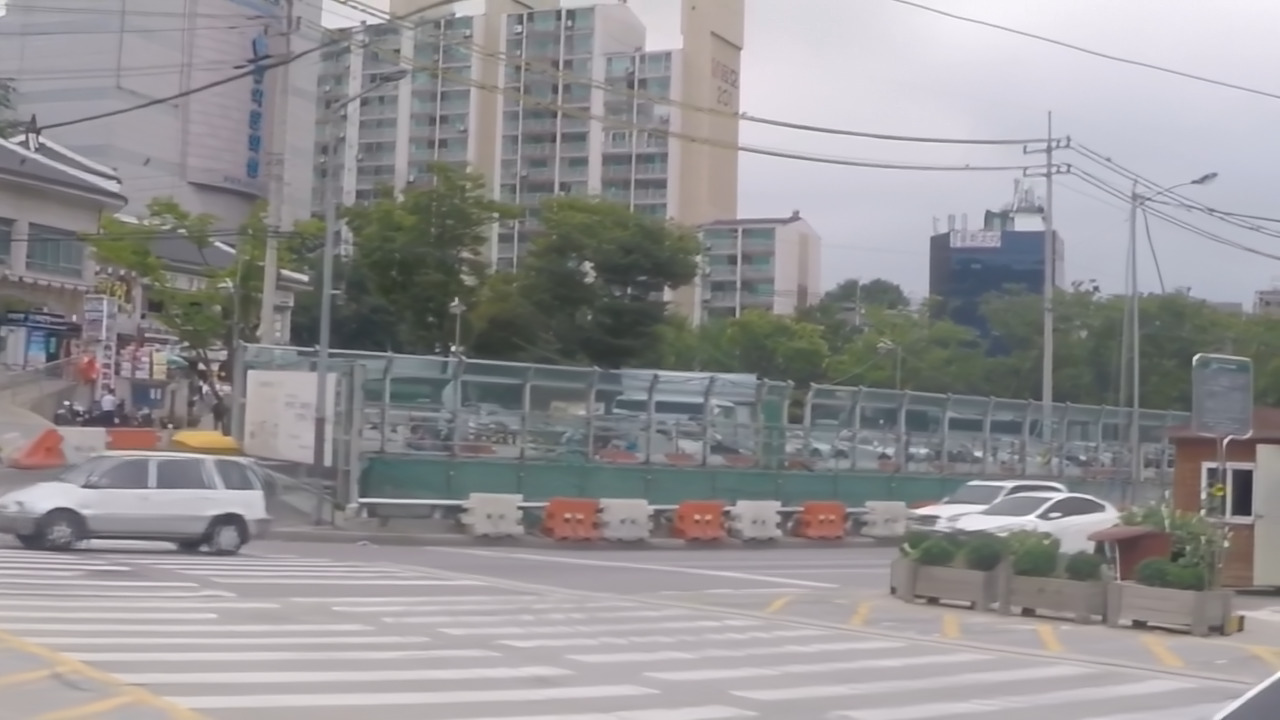}								
\\
				\includegraphics[width=0.12\textwidth, height=0.085\textwidth]{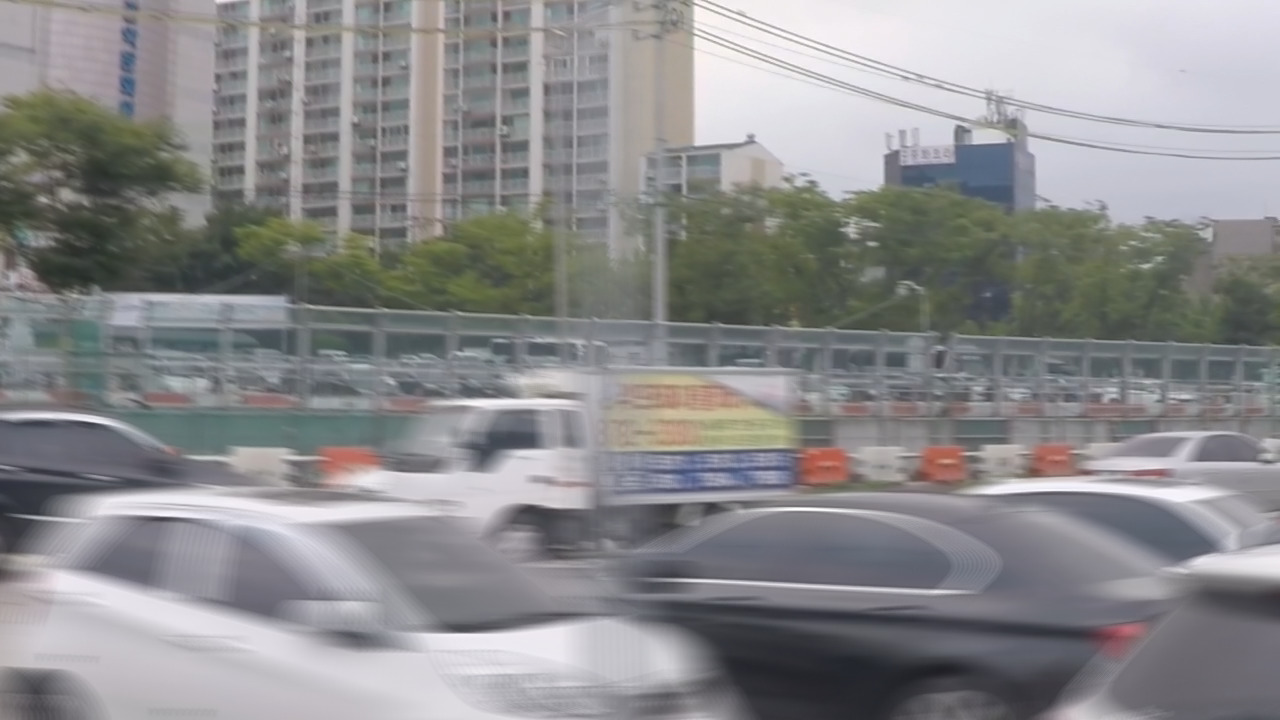}
				&
				\includegraphics[bb=590 240 720 340,clip=True,width=\widthscaleblur \textwidth]{deblurring2/000209_blur} & 

				\includegraphics[bb=590 230 720 330,clip=True,width=\widthscaleblur \textwidth]{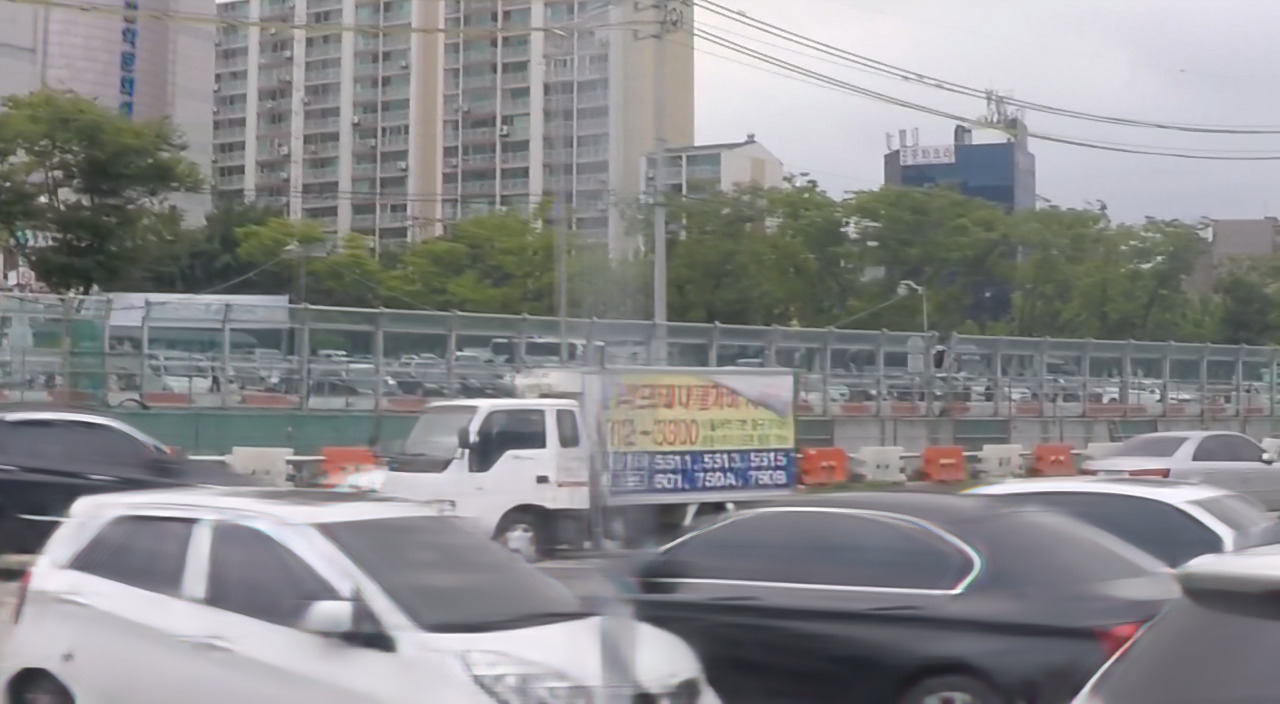} & 
				\includegraphics[bb=590 240 720 340,clip=True,width=\widthscaleblur \textwidth]{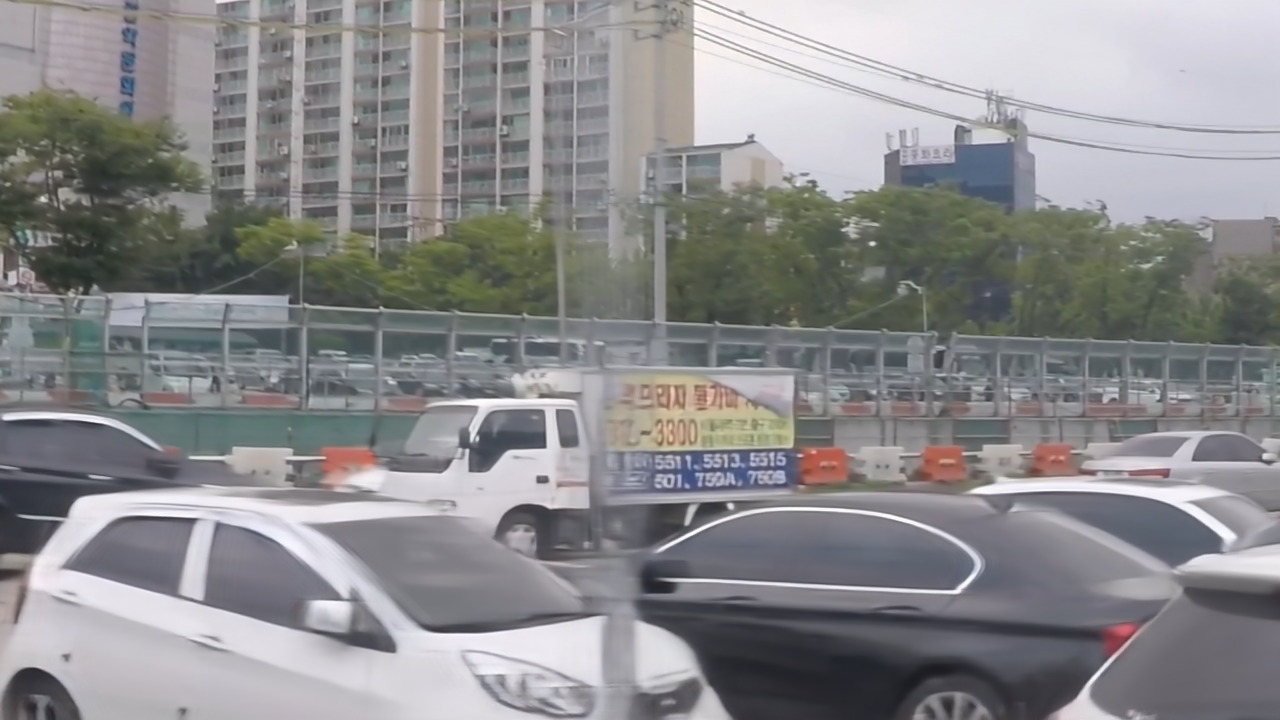} &
				\includegraphics[bb=590 240 720 340,clip=True,width=\widthscaleblur \textwidth]{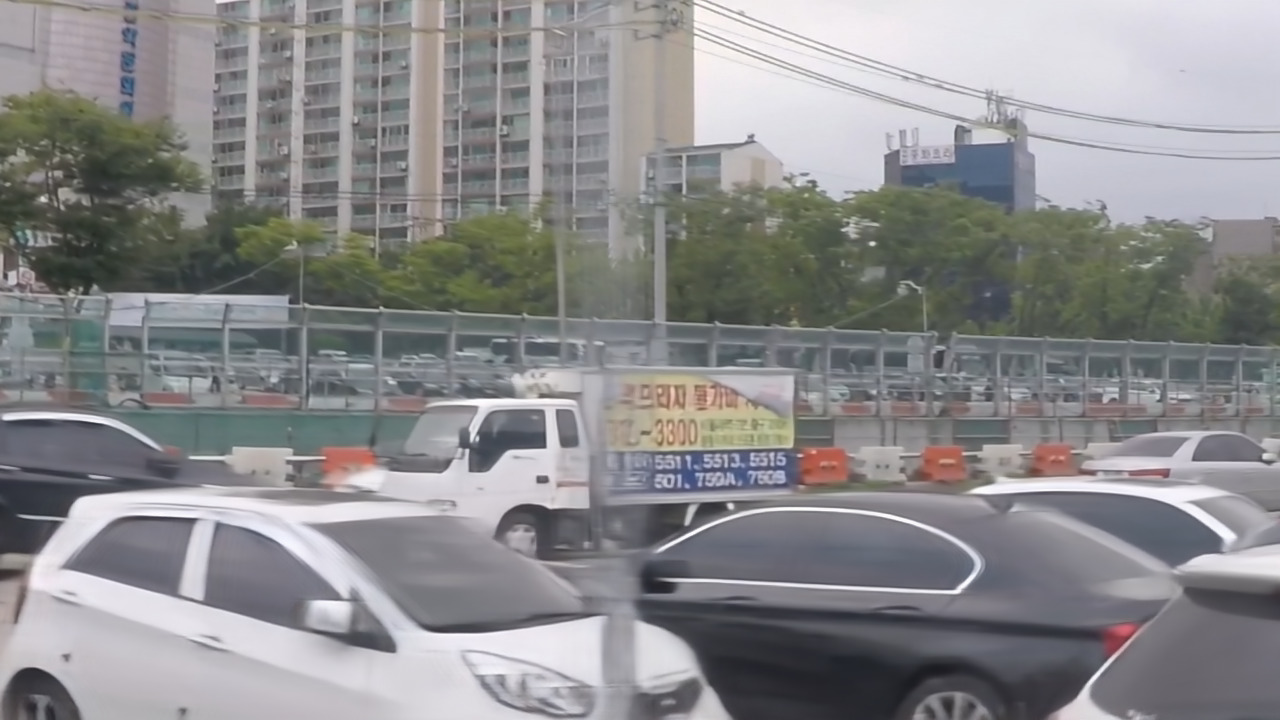} &
				\includegraphics[bb=590 240 720 340,clip=True,width=\widthscaleblur \textwidth]{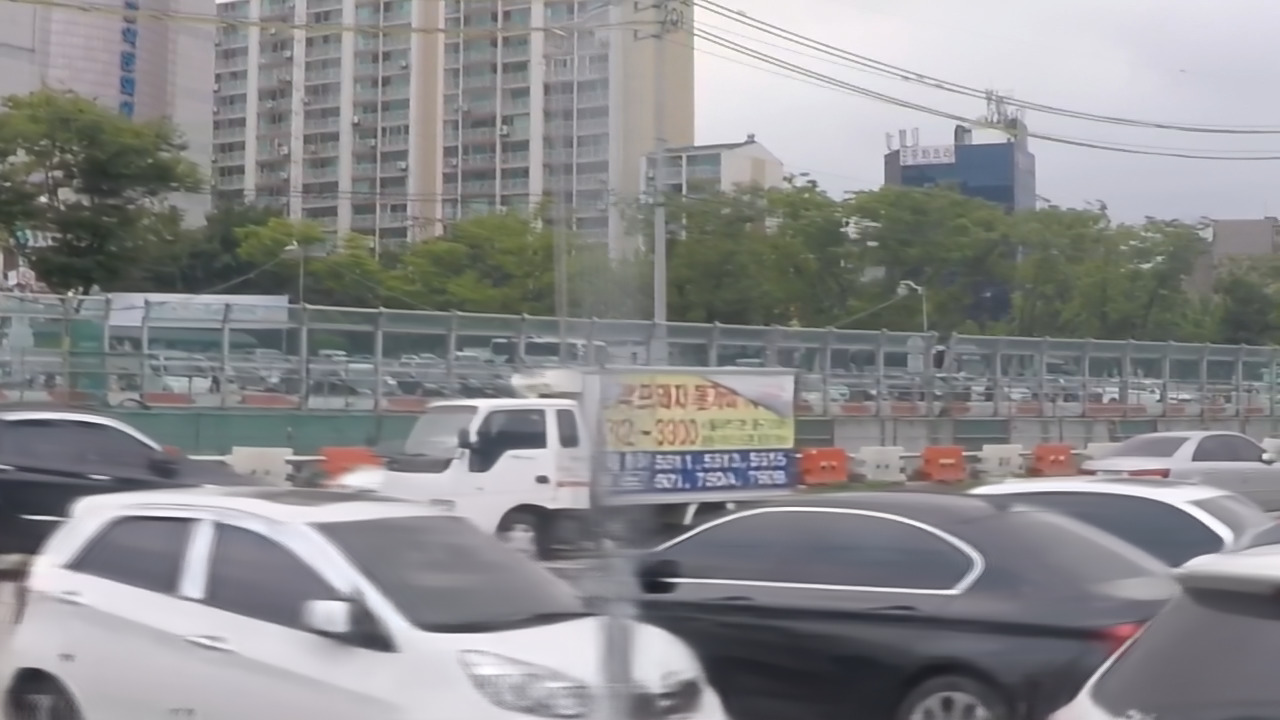} &				 
				\includegraphics[bb=590 240 720 340,clip=True,width=\widthscaleblur \textwidth]{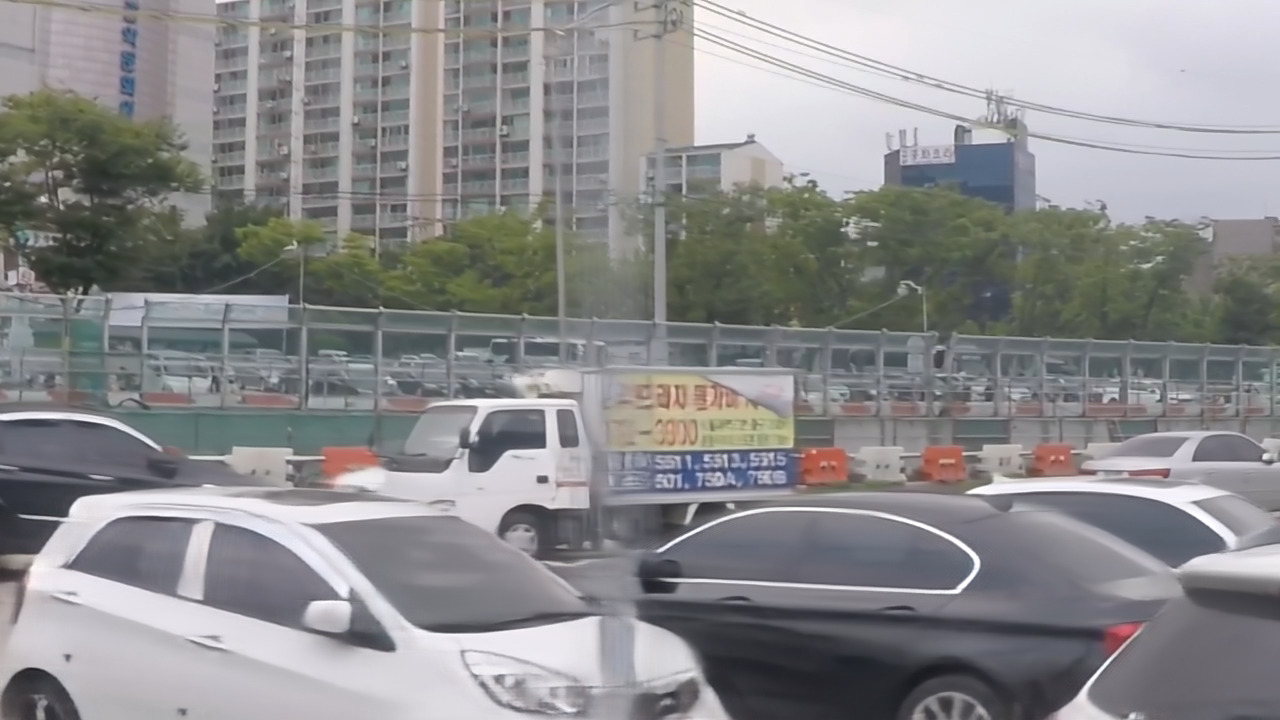} &
				\includegraphics[bb=590 240 720 340,clip=True,width=\widthscaleblur \textwidth]{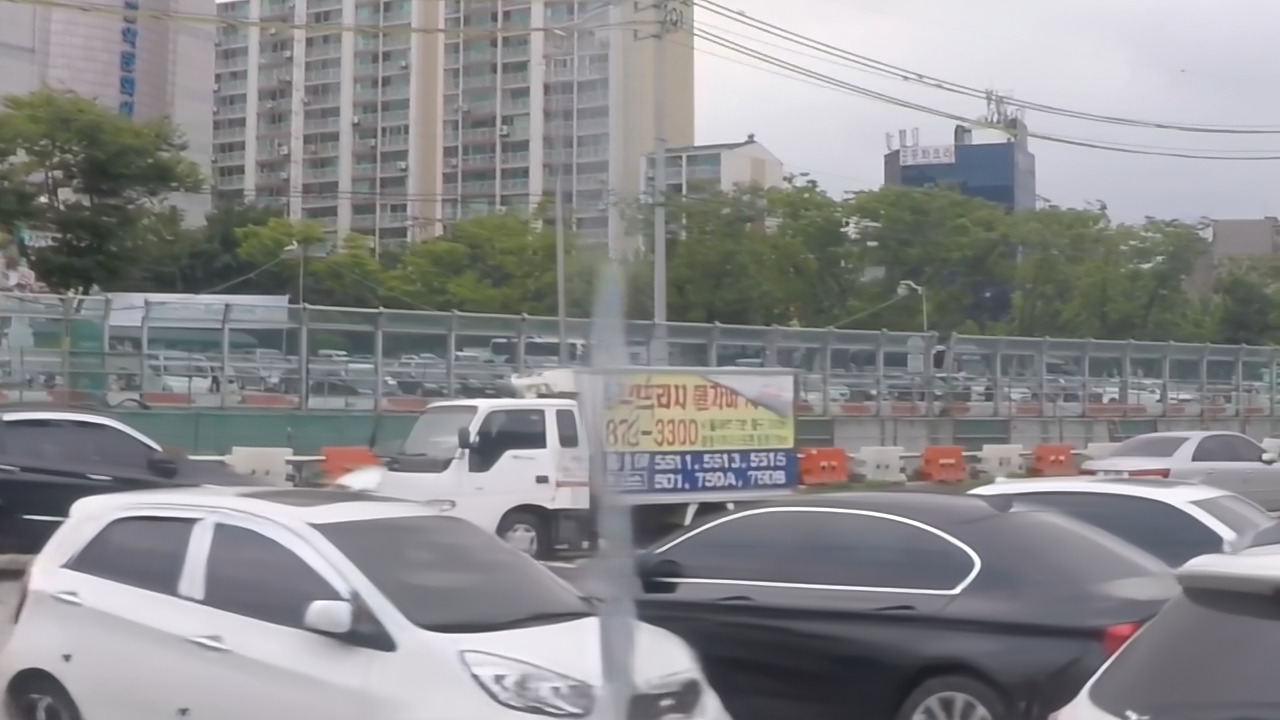}				
				 
				\\ 							
								 Blurred Image&

				 Blurred patch& 
				 SRN & 
				 DelurGAN-V2& 
				 \scriptsize{Stack(4)-DMPHN} & 
				 MTRNN & 
				 Suin et al. & 
				 \ourmethod{}
				\\
	\end{tabular}
	\caption{Visual comparisons of zoomed-in results of competing deblurring models on images from the GoPro test set~\cite{nah2017deep}.} 
\label{fig:dynamic}
 \vspace{-0.4cm}
\end{figure*}

Visual comparisons on images containing dynamic and 3D scenes are shown in Fig. ~\ref{fig:dynamic}. Often, the results of prior works suffer from incomplete deblurring or artifacts. In contrast, our network demonstrates non-uniform deblurring capability while preserving sharpness. Scene details in the regions containing text, boundaries, and textures are more faithfully restored, making them recognizable.

%% file: paper_parts/ablation2.tex
\begin{table}[t]
\caption{Network analysis with PSNR for AGAN and GoPro benchmarks, respectively. DED, SFM, SC, NL, SNL denote dense encoder-decoder, spatial feature modulator, mask-guided sparse convolution, non-local module, and mask-guided sparse non-local module, respectively.\label{table:ablation}}
\resizebox{0.98\linewidth}{!}{%
\begin{tabular}{@{}ccccccccc@{}}
\toprule
Methods & DED & SFM & SC & NL & SNL & \multicolumn{2}{c}{PSNR} \\
& & & & & & Raindrop & Motion Blur\\
\midrule
Net1 & \checkmark &  &  &  &  & 30.72 & 30.85 \\
Net2 & \checkmark  & \checkmark &  &  &   & 31.39 & 31.44 \\
Net3 & \checkmark  & \checkmark & \checkmark &  &  & 31.80 & 31.62 \\
Net4 & \checkmark & \checkmark & \checkmark & \checkmark &   & 32.17 & 31.79 \\
\textbf{\ourmethod{}} & \checkmark & \checkmark & \checkmark &  &  \checkmark & \bf{32.73} & \bf{32.06} \\
\bottomrule
\end{tabular}%
  \label{table:ablation}
}
 \vspace{-0.6cm}
\end{table}

This work explores the benefits of distortion-localization guided feature modulation and sparse processing for spatially-varying restoration tasks. Table \ref{table:ablation} quantifies the effect of individual design choices on performance of \ourmethod{} on the AGAN (raindrop) and GoPro (motion blur) datasets. 

To validate our design choices, we implement the following
baselines (reported in Table \ref{table:ablation}). Net1: Dense encoder-decoder network (CNN backbone of our $Net_R$) with few additional parameters to match $Net_L$. Net2: Net1 guided by $Net_L$ using SFM. Net3: Net2 with all densely connected convolutional blocks in decoder replaced with SC modules, Net4: Net3 with non-local (NL) layer \cite{wang2018non} introduced in the decoder. Net5: Net4 containing the proposed SNL module instead of NL. Good baseline scores of Net1 for both tasks support our backbone design choice.

\noindent \textbf{Effectivenes of SFM:} Net2 introduces SFM blocks (Sec. \ref{sec:knowledge_transfer}) which guide restoration network using mask and features of $Net_L$ at multiple intermediate levels. The significant improvement in accuracy in comparison to Net1 demonstrates the benefit of degradation guidance. 

\noindent\textbf{Effect of SC and SNL modules:} Net4 employs the general non-local layer \cite{wang2018non} in decoder global context aggregation. 
Net5 has the same structure as Net4 (sans the NL module), and it feeds the predicted mask as input to the SNL modules. 
The improvement in behavior and performance is attributed to SNL design which uses explicit distortion-guidance to steer pixel-attention. SNL is more suited than NL for both degraded and clean regions. As explained in Eq. \ref{eqn:pairwise_relationship2}, while restoring degraded pixels, SNL assigns dynamically estimated non-zero weights to features originating from only clean pixels in the image. By design, it leaves the features of clean regions unaltered. 
As reported in Table \ref{table:ablation}, Net3 vs. Net2 shows the benefits of SC module whereas, Net5 vs. Net3 shows the utility of global context aggregation for restoration. Net5, our final model, shows a significant improvement over CNN baseline (Net1), demonstrating the advantages of our overall solution over static CNNs.

\noindent\textbf{Supplementary Details:} We provide additional real results and qualitative comparisons for all four tasks, additional model analysis, and results of multi-task learning in the supplementary document.

\noindent\textbf{Benefit:} Many applications (e.g., autonomous vehicles) involve dealing with rain, shadows, blur etc. at different time instances. Designing architectures that are applicable across multiple tasks, without requiring specialized architecture re-engineering is practically very convenient 
(potentially facilitating customized hardware design). Our versatile design enables this as only the learned weights vary across degradations while the architecture remains the same.
\vspace{-0.3cm}

%% file: paper_parts/conclusion.tex
\vspace{-0.2cm}
We addressed the single image restoration tasks of removing spatially-varying degradations such as raindrop, rain streak, shadow, and motion blur. We model the restoration
task as a combination of degraded-region localization
and region guided sparse restoration and propose a guided
image restoration framework \ourmethod{} which leverages
the features of $Net_L$ for spatial modulation of the features in $Net_R$ using SFM module. We introduce distortion localization awareness in $Net_R$ using sparse convolution module (SC) and sparse
non-local attention module (SNL) and show its significant
benefits . Extensive evaluation on $11$ datasets across four restoration tasks demonstrates that proposed framework outperforms strong degradation-specific baselines. Ablation analysis and visualizations are shown to validate the effectiveness of its key components.